%% file: Arxiv.tex
\documentclass[11pt,twoside]{article} 

\usepackage{eqnarray,amsmath}
\usepackage[utf8]{inputenc} 
\usepackage[T1]{fontenc}    
\usepackage{booktabs}       
\usepackage{amsfonts}       
\usepackage{nicefrac}       
\usepackage{microtype}      

\usepackage{epsf}
\usepackage{epsfig}
\usepackage{fancyhdr}
\usepackage{graphics}
\usepackage{graphicx}
\usepackage{psfrag}
\usepackage{fullpage}
\usepackage{pdfpages}
\usepackage{subcaption}

\usepackage{url}
\usepackage[colorlinks,linkcolor=magenta,citecolor=blue, pagebackref=true]{hyperref}
\renewcommand*{\backrefalt}[4]{%
    \ifcase #1 \footnotesize{(Not cited.)}%
    \or        \footnotesize{(Cited on page~#2.)}%
    \else      \footnotesize{(Cited on pages~#2.)}%
    \fi}

\usepackage{color}

\usepackage{amsthm}
\usepackage{amsmath}
\usepackage{amssymb,bbm}
\usepackage{caption}
\usepackage{textcomp}
\usepackage{siunitx}
\usepackage{wrapfig}
\usepackage{multirow}
\usepackage{multicol}

\input{macro_commands}

\begin{document}

\begin{center}

{\bf{\LARGE{Attack On Prompt: \\
Backdoor Attack in Prompt-Based Continual Learning}}}
  
\vspace*{.2in}
{\large{
\begin{tabular}{ccc}
Trang Nguyen$^{\diamond}$ & Anh Tran $^{\diamond}$ & Nhat Ho$^{\dagger}$
\end{tabular}
}}

\vspace*{.2in}

\begin{tabular}{c}
The University of Texas at Austin$^{\dagger}$\\
VinAI Research$^{\diamond}$
\end{tabular}

\vspace*{.1in}
\today

\vspace*{.2in}

\end{center}

\begin{abstract}
Prompt-based approaches offer a cutting-edge solution to data privacy issues in continual learning, particularly in scenarios involving multiple data suppliers where long-term storage of private user data is prohibited. Despite delivering state-of-the-art performance, its impressive remembering capability can become a double-edged sword, raising security concerns as it might inadvertently retain poisoned knowledge injected during learning from private user data. Following this insight, in this paper, we expose continual learning to a potential threat: backdoor attack, which drives the model to follow a desired adversarial target whenever a specific trigger is present while still performing normally on clean samples. We highlight three critical challenges in executing backdoor attacks on incremental learners and propose corresponding solutions: (1) \emph{Transferability}: We employ a surrogate dataset and manipulate prompt selection to transfer backdoor knowledge to data from other suppliers; (2) \emph{Resiliency}: We simulate static and dynamic states of the victim to ensure the backdoor trigger remains robust during intense incremental learning processes; and (3) \emph{Authenticity}: We apply binary cross-entropy loss as an anti-cheating factor to prevent the backdoor trigger from devolving into adversarial noise. Extensive experiments across various benchmark datasets and continual learners validate our continual backdoor framework, with further ablation studies confirming our contributions' effectiveness.
\end{abstract}

%

\section{Introduction}
\label{introduction}

The adaptability of human learning to absorb new knowledge without forgetting previously acquired information remains a significant challenge for machine learning models. Continual learning (CL) endeavors to narrow this chasm by guiding models to sequentially learn new tasks while maintaining high performance on earlier ones. An outstanding solution to CL is the prompt-based approach \cite{smith2023codaprompt, wang2022dualprompt, wang2022learning, wang2023hide, qiao2024PGP}, which leverages the power of pre-trained models and employs a set of trainable prompts for flexible model instruction, accommodating data from various tasks. Thanks to its ability to remember without storing a memory buffer, prompt-based CL methods are particularly suitable for scenarios prioritizing data privacy, such as those involving multiple data suppliers.

\begin{figure}[!t]

  \centering
  \includegraphics[width=0.8\textwidth]{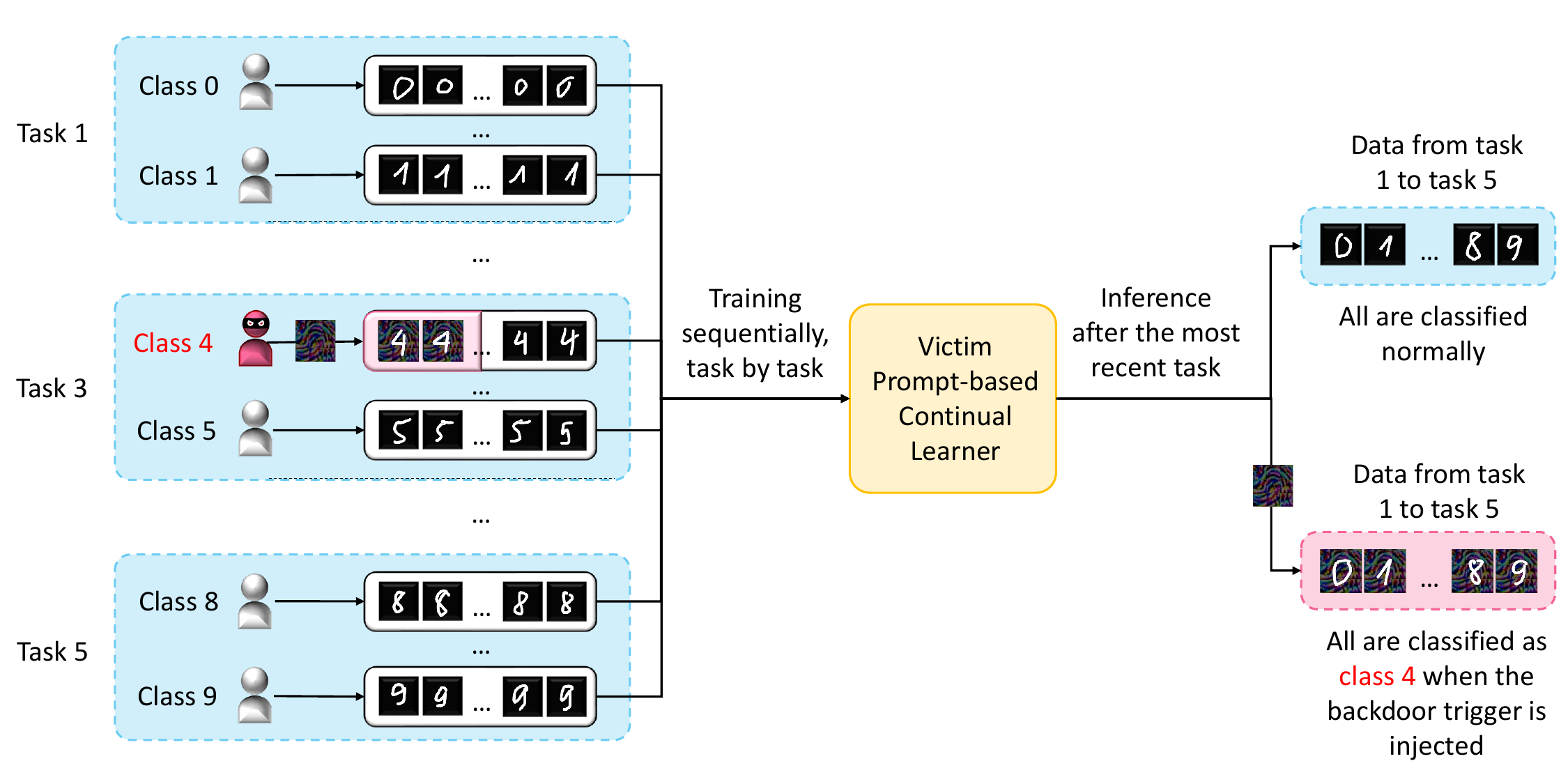}
  \caption{Multi data supplier scenario in prompt-based continual learning, with one supplier acting as an adversarial attacker.}
  \label{figure: threatmodel}
\end{figure}

Nonetheless, such promising results can inadvertently become vulnerabilities, exposing CL to security threats. Indeed, while CL methods effectively address catastrophic forgetting by preserving and incorporating previously acquired knowledge, they may also unwittingly retain knowledge compromised by adversarial actions. These threats become even more formidable in the multi-data supplier scenario of prompt-based approaches, where the supplied data might contain hidden harmful information, as shown in Figure \ref{figure: threatmodel}.

One potential threat is backdoor attack, which manipulates neural networks to exhibit the attacker’s desired behavior when the input contains a specific backdoor trigger. Typically, adversaries poison a small portion of the training data, causing models trained on this data to misclassify any images with the triggers as a given target class while performing normally on clean samples. This makes the attack less likely to be suspected by the victim learner. As backdoor attacks pose such dangerous threats, increasingly sophisticated methods are being introduced. These include black-box scenarios where the attacker has no information about the model and learning procedure \cite{saha2019hidden, souri2021sleeper, turner2019labelconsistent}, or data-constrained cases where adversaries control only a fragment of the training data \cite{zeng2022narcissus, li2024efficient}. With high efficacy, even in these challenging situations, backdoor attacks are particularly threatening in multi-data supplier scenarios. In spite of significant attention in various tasks and areas such as computer vision \cite{turner2019labelconsistent, liao2018backdoor, moosavidezfooli2017universal, 9709953, NEURIPS2021_9d99197e, nguyen2021wanet}, large language models and natural language processing \cite{cai2022badprompt, li2024badedit}, point clouds \cite{xiang2021backdoor, xiang2019generating, li2021pointba}, federated learning \cite{Xie2020DBA, wang2020attack, zhang2022neurotoxin, dai2023chameleon}, and more, targeted black-box backdoor attacks have not been thoroughly explored in continual learning.

\textbf{Challenges.} Despite holding such potential danger for CL, extending backdoor attacks to the incremental setting is non-trivial. Firstly, in the multi-supplier setting where the victim gathers data from different sources, the attacker lacks information about the actual data distribution used to train the victim model. Consequently, \emph{generalizing backdoor knowledge to be transferable to unknown data} poses the first challenge that our continual backdoor approach must confront. The second challenge arises from the vulnerability of backdoor attacks during fine-tuning. Recent studies \cite{sha2022finetuning, min2023towards} have highlighted the tendency for backdoor knowledge to be removed when the victim fine-tunes the poisoned model on a small and clean dataset. This issue is exacerbated in continual learning, where the \emph{victim model undergoes incremental training} as new data from various sources arrive. The final challenge involves the backdoor trigger's proneness to turn into adversarial noise. Huynh et al. \cite{Huynh_Nguyen_Pham_Tran_2024} observed that the trigger, when optimized using a surrogate model, may \emph{transform into an adversarial perturbation}, driving the clean model to follow desired adversarial targets even in the absence of any prior backdoor attacks. Since conventional adversarial defenses can mitigate such adversarial noise, preempting this behavior is crucial to strengthen the resilience of the backdoor trigger.

\textbf{Contributions.} In response to these shortcomings, we propose a continual backdoor framework that satisfies three key properties: \emph{\textbf{transferability} to unknown data}, \emph{\textbf{resilience} to incremental learning procedures}, and \emph{\textbf{authenticity} to avoid becoming adversarial noise}. Initially, we leverage the natural label mapping characteristic of visual prompting, thereby approaching the data poisoning issue from the perspective of prompt selection. This approach allows our backdoor trigger to be generalized to any victim data distribution. Next, we robustify the backdoor trigger by aligning the optimization process with the continuously changing states of the incremental learner, thus ensuring the effectiveness of the backdoor trigger when the model is trained on new incoming clean data. Finally, we reconsider the choice of loss function for trigger optimization. We observe that the commonly used softmax function with cross-entropy introduces bias towards the target class, pushing its score excessively high and leading to the adversarial noise problem. Building on this observation, we propose adopting binary cross-entropy (BCE) with sigmoid function to mitigate this issue, thereby eliminating the dependency of trigger optimization on other classes and preventing cheating behavior.

By integrating the components above, our framework, termed backdoor-\textbf{A}ttack \textbf{O}n \textbf{P}rompt-based CL (\textbf{AOP}), successfully backdoor-attacks continual learners, achieving an Attack Success Rate (ASR) of up to $100\%$. Our contributions are three-fold and can be summarized as follows:

\textbf{1.} We expose prompt-based CL to backdoor attacks. Our approach follows strong assumptions, with black-box, clean-label, and constrained-data setting;

\textbf{2.} We highlight three key challenges that our continual backdoor framework must address: ensuring transferability to unknown data in prompt tuning, preventing the catastrophic forgetting of backdoor knowledge, and mitigating the tendency to generate adversarial noise due to biases. 

Motivated by these challenges, we propose a novel continual backdoor framework comprising three main components: utilizing a surrogate dataset to manipulate prompt selection, dynamically optimizing the backdoor trigger, and adopting sigmoid BCE loss to mitigate bias and prevent cheating;

\textbf{3.} We conduct extensive experiments on various prompt-based continual learners with different datasets and provide ablation studies to demonstrate the strength of our framework.

\section{Background}
\label{section: preliminaries}
\textbf{Continual learning.} 
In continual learning scenarios, the model undergoes a sequential presentation of tasks ${\mathcal{D}_{1},... \mathcal{D}_{T}}$. Each task corresponds to distinct subsets of tuples $\mathcal{D}_{t} = \{\boldsymbol{x}_{t}^{i}, \boldsymbol{y}_{t}^{i}\}_{i=1}^{i=n_{t}}$, where $\boldsymbol{x}_{t}^{i} \in \mathcal{X}^{t}$ is the input sample, $\boldsymbol{y}_{t}^{i} \in \mathcal{Y}^{t}$ is the corresponding label, and $n_{t}$ is the number of samples for task $t$. It is important to note that data from prior tasks become inaccessible during the training of subsequent tasks \cite{smith2023codaprompt, qiao2024PGP}. The objective of continual learning is to continuously acquire the capability to classify newly introduced classes while maintaining proficiency on previously learned ones in a single model $f: \mathcal{X} \rightarrow \mathcal{Y}$. In this paper, and in prompt-based methods \cite{smith2023codaprompt, wang2022dualprompt, wang2022learning, wang2023hide, qiao2024PGP}, $f$ represents the pre-trained Vision Transformer (ViT) encoder. Additionally, $\phi$ is employed as the shared classification head, and $\phi_{t}$ is the classifier corresponding to classes specific to the given task $t$.


\textbf{Prompt-based continual learning.} We provide a concise overview of L2P \cite{wang2022learning}, which stands as the first work that integrates prompts into the context of continual learning. L2P introduces a prompt pool comprising learnable prompts and their corresponding keys $\left\{\left(\boldsymbol{k}_{1}, \boldsymbol{p}_{1}\right),\left(\boldsymbol{k}_{2}, \boldsymbol{p}_{2}\right), \cdots,\left(\boldsymbol{k}_{n_p}, \boldsymbol{p}_{n_p}\right)\right\}$ where $n_p$ is total number of prompts. These prompts are then combined with image features and fed into the pre-trained ViT $f$, instructing the model to perform classification. Prompts are queried in an instance-wise manner using the top-$K$ cosine similarity $\gamma\left(q(\boldsymbol{x}), \boldsymbol{k}_{i}\right)$ between the keys and the query function $q(\boldsymbol{x})=f(\boldsymbol{x})[0,:]$. Subsequent prompt-based methods are designed based on L2P, each featuring prompt utility and optimization modifications. A brief explanation of these methods can be found in the supplementary material.

\section{Backdoor Attack on Prompt-based Continual Learning (AOP)}
\label{section: method}
\label{method}

We first outline the threat model and introduce key notations in Section~\ref{method:setting}. We then delineate the three primary components of AOP across Sections~\ref{method:transferability}-\ref{method:bce}. A comprehensive overview and the end-to-end algorithm is in the supplementary material.

\subsection{Threat Model and Notations}
\label{method:setting}

\textbf{Continual learning protocols.} We consider the class-incremental learning (CIL) setting in prompt-based continual learning \cite{wang2022dualprompt, wang2022learning}. In CIL, training data for incremental tasks $\mathcal{D}_t$ arrive incrementally in a discrete manner. Each task consists of data for new $M$ classes that have not been learned by the model before.  Formally, each task $\mathcal{D}_t = {\{\mathcal{D}_{m, t}\}_{m=1}^{M}}$ with each class $\mathcal{D}_{m, t} = {\{\boldsymbol{x}_{i}^{m, t}, y_{i}^{m, t}\}}^{n_{m, t}}_{i=1}$ comprises input samples $\boldsymbol{x}_{i}^{m, t} \in \mathcal{X}$ and their corresponding labels $y_{i}^{m, t} = c_{m, t} \in \mathcal{Y}$, where $n_{m, t}$ represents the number of training samples for the corresponding class. In CIL, the learner is required to perform classification across all classes encountered up to task $T$ without being provided with explicit task labels during inference. Data for different classes $m$ and $m^{\prime}$ are gathered from different suppliers. To ease the ensuing presentation, the index $t$ is omitted unless noted otherwise.

\textbf{Backdoor attack protocols.} Let the attacker be the data supplier for class $m$ with labels $c_m$. The attacker's goal is to poison the supplying dataset with a small amount of trigger-injected samples, such that any data from any classes if manipulated with the backdoor trigger, will be misclassified as $c_m$ by the resulting incremental victim model when performing inference at any time $t$.  An example of a triggered image is given in the supplementary material.

Consider $\mathcal{D}_m =\left\{\left(\boldsymbol{x}_{i}, y_{i}\right)\right\}_{i=1}^{n_m}$ as the benign training set of class $m$. The adversary then learns to generate the poisoned dataset $\mathcal{D}_p$. Specifically, $\mathcal{D}_p$ consists of two parts: a modified version of a selected subset (denoted as $\mathcal{D}_s$) of $\mathcal{D}_m$ and the remaining benign samples. Thus, $\mathcal{D}_p=\mathcal{D}_c \cup \mathcal{D}_b$, $\mathcal{D}_c=\mathcal{D}_m \backslash \mathcal{D}_s$, $\mathcal{D}_b=\left\{\left(\boldsymbol{x}^{\prime}, c_m\right) \mid \boldsymbol{x}^{\prime}=G(\boldsymbol{x}),(\boldsymbol{x}, c_m) \in \mathcal{D}_s\right\}$, where $c_m$ is the adversary target label, $\gamma \triangleq \frac{\left|\mathcal{D}_s\right|}{|\mathcal{D}_m|}$ is the poisoning rate, and $G: \mathcal{X} \rightarrow \mathcal{X}$ is an adversary-specified poisoned image generator. We follow \cite{souri2021sleeper,  li2021invisible} and formulate $G(\boldsymbol{x})=\boldsymbol{x}+\boldsymbol{\delta}$, where the perturbation $\boldsymbol{\delta}$ has a bounded $\ell_{p}$-norm.

We emphasize that given the considered multi-data supplier scenario, we optimize the backdoor trigger following a \emph{black-box} setting (where the attacker has no access to the training model or procedure) and a \emph{clean-label} setting (where the attacker cannot change the label of data), which represent stealthy and challenging conditions in backdoor attacks. 

\subsection{Prompt Selection, Label Mapping, and Transferability}
\label{method:transferability}

\begin{figure}[t!]
     \centering
     \begin{subfigure}[b]{0.3\textwidth}
         \centering
         \includegraphics[width=\textwidth]{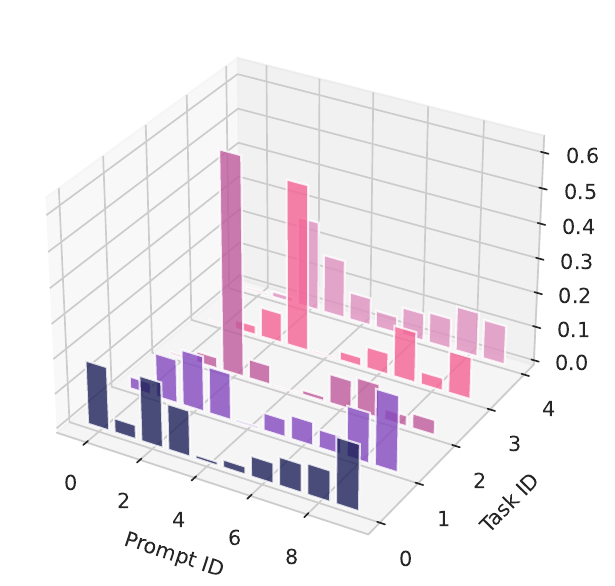}
         \caption{Benign samples}
         \label{fig:clean_frequency_task}
     \end{subfigure}
     \hfill
     \begin{subfigure}[b]{0.3\textwidth}
         \centering
         \includegraphics[width=\textwidth]{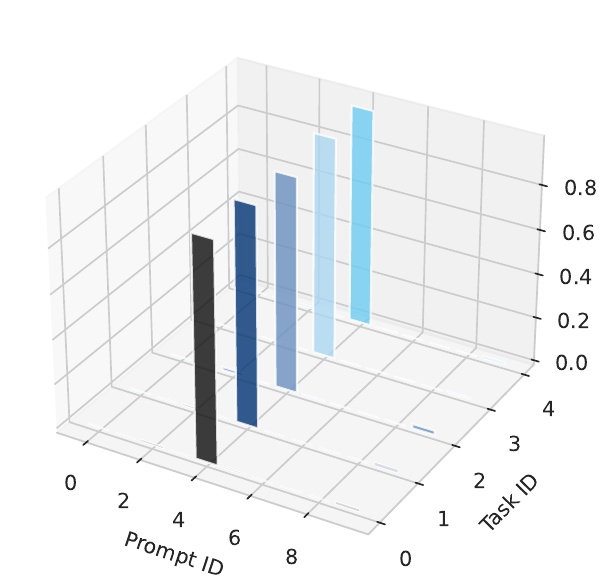}
         \caption{Triggered samples}
         \label{fig:poison_frequency_task}
     \end{subfigure}
     \begin{subfigure}[b]{0.3\textwidth}
         \centering
         \includegraphics[width=\textwidth]{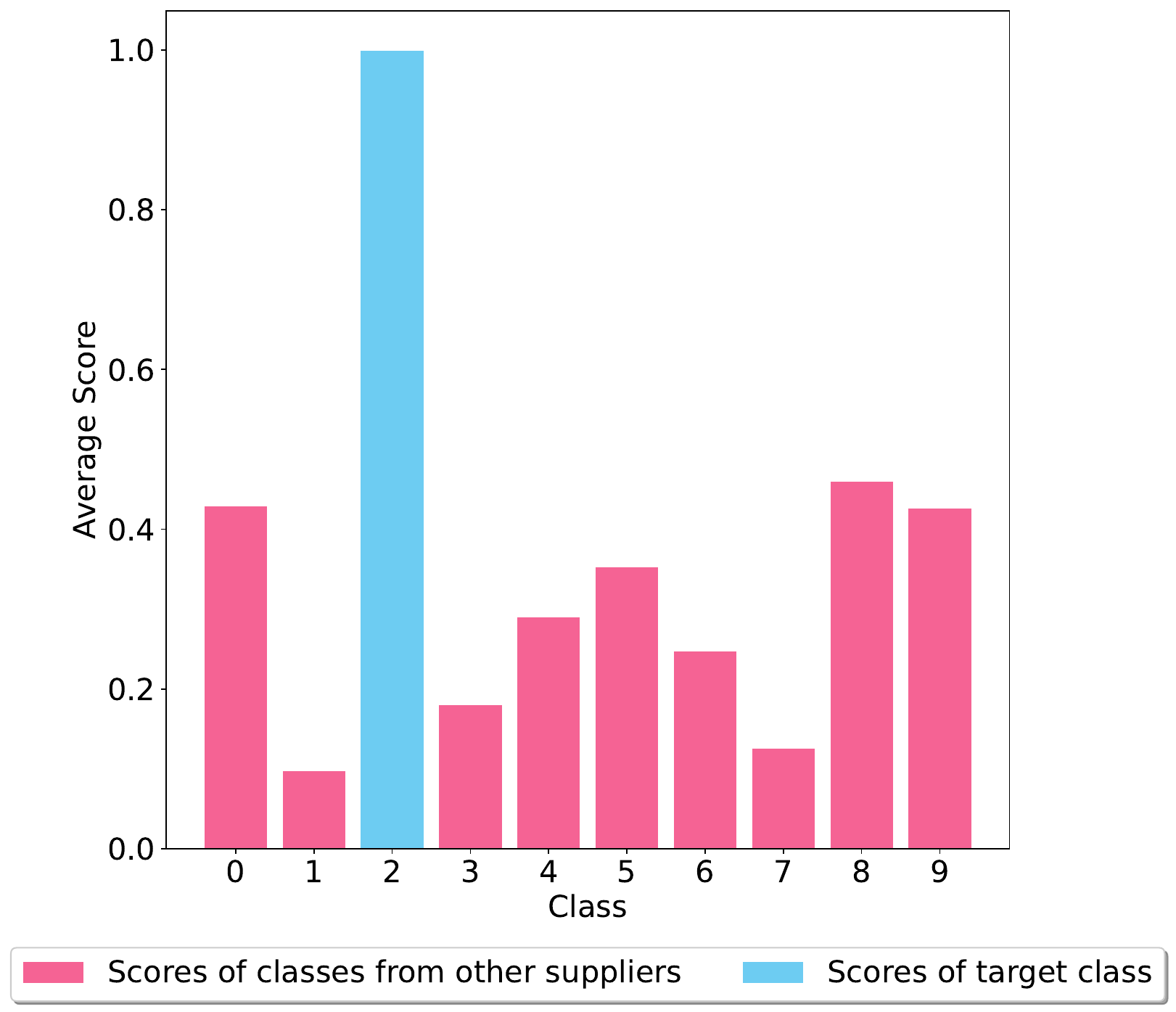}
         \caption{Cheated trigger}
         \label{fig:cheat_average_logits}
     \end{subfigure}
     \hfill
     \begin{subfigure}[b]{0.3\textwidth}
         \centering
         \includegraphics[width=\textwidth]{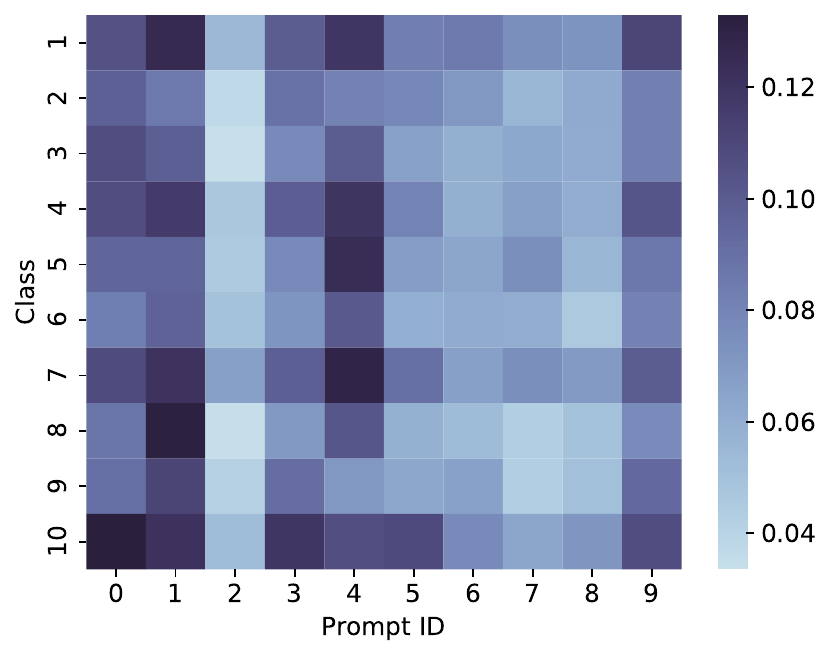}
         \caption{Benign samples}
         \label{fig:clean_correl}
    \end{subfigure}
    \hfill
     \begin{subfigure}[b]{0.3\textwidth}
         \centering
         \includegraphics[width=\textwidth]{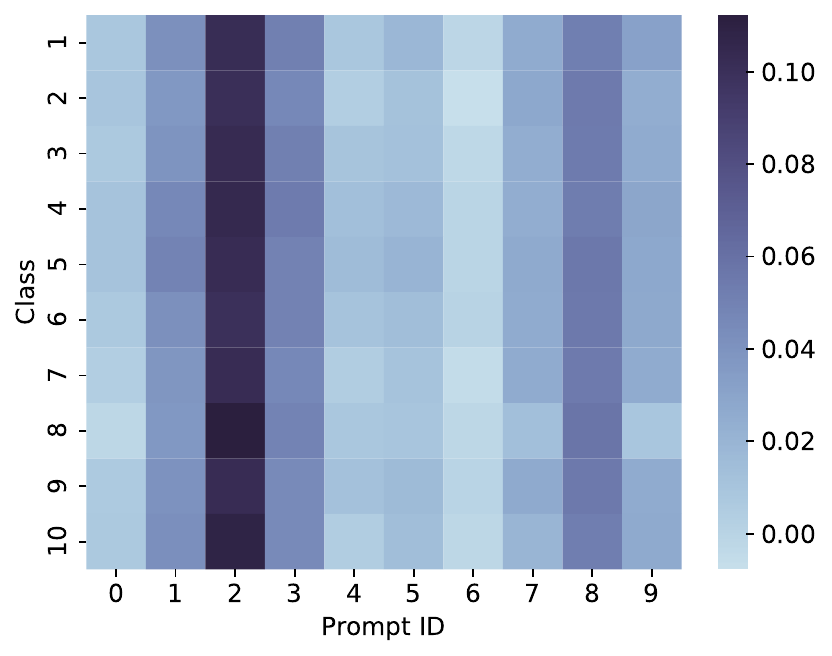}
         \caption{Triggered samples}
         \label{fig:ba_correl}
    \end{subfigure}
    \hfill
     \begin{subfigure}[b]{0.3\textwidth}
         \centering
         \includegraphics[width=\textwidth]{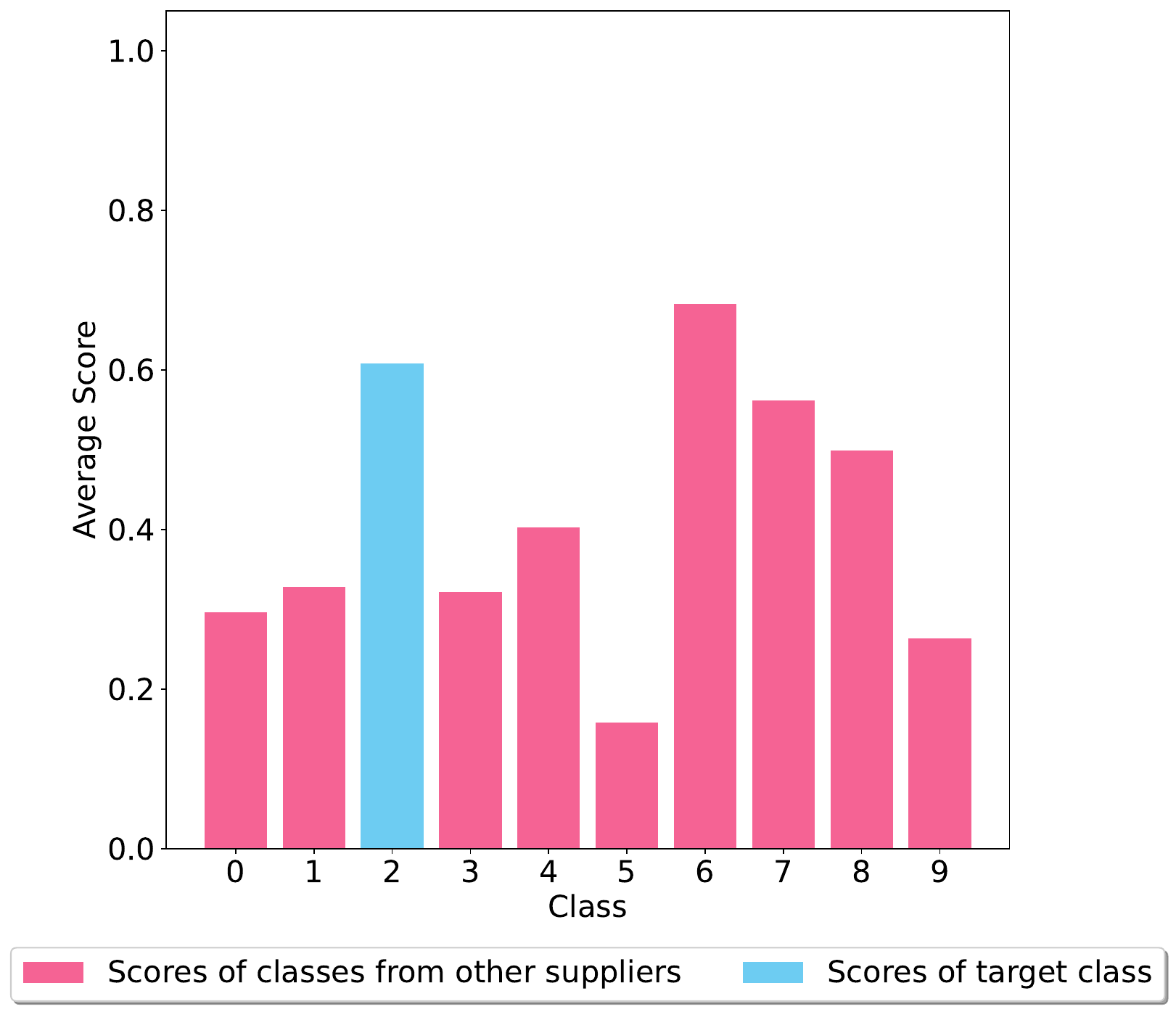}
         \caption{Authentic trigger }
         \label{fig:bce_average_logits}
     \end{subfigure}
     \hfill
        \caption{(a) and (b): AOP's prompt selection frequency on benign and triggered samples when attacking DualPrompt. (d) and (e): AOP's average key-query similarities concerning benign and triggered samples when attacking DualPrompt-PGP. (c) and (f): Scores obtained from the clean model for AOP's triggered samples optimized with CE and BCE, respectively.}
        \label{fig:main_viz}
        \vspace{-0.8 em}
\end{figure}

The core of prompt-based continual learning methods lies in the prompt pool and the prompt selection strategy. Specifically, the most relevant prompts are queried in an instance-wise manner and then concatenated with the sample to optimally guide the model in performing classification. We leverage this fundamental mechanism of the prompt-based approach to reframe the backdooring problem as one of manipulating prompt selections. As in Figures~\ref{fig:clean_frequency_task} and~\ref{fig:poison_frequency_task}, we aim to ensure that triggered samples are directed to select specific backdoor prompts, thereby causing the model to misclassify these backdoor-prompted samples into the desired class.

A key feature of visual prompting is its ability to act as a label mapping mechanism when performing downstream tasks using a pretrained model. In this context, prompts function as universal input perturbation templates, enabling the mapping of labels from a source dataset to a target dataset \cite{Chen_2023_CVPR}. From this perspective, our aim of controlling prompt selection translates into manipulating label mappings between the two datasets. This new perspective paves the way for the "transferability" of our continual backdoor framework.

When optimizing the backdoor trigger, we employ a surrogate dataset, denoted as $\mathcal{D}_\text{surrogate}$, to address the backdoor transferability to data from other classes. It is worth noting that $\mathcal{D}_\text{surrogate}$ does not necessarily mirror the actual data distribution used to train the incremental model. This discrepancy stems from the visual prompting property discussed earlier. In particular, instead of optimizing a trigger that causes the poisoned data to be misclassified by the model, our backdoor trigger can be viewed as activating an incorrect mapping to the target class. Since we focus on manipulating the mapping and prompt selection rather than the dataset itself, $\mathcal{D}_\text{surrogate}$ can be chosen differently from the actual dataset to align with our objectives.

\subsection{Static-dynamic Trigger Optimization}
\label{method:static-dynamic}

Since we lack information about the victim's continual model, we use $\mathcal{D}_\text{surrogate}$ to train a surrogate incremental learner. We then optimize the backdoor trigger $\boldsymbol{\delta}$ based on this surrogate incremental model. Specifically, we employ the surrogate learner with two states: a static state that reflects how prompts learn label mappings between the source and target datasets, and a dynamic state that reflects the continuous learning procedure of the victim model. Formally, our static-dynamic trigger optimization involves the following four stages:

\textbf{(0) Preparation} 
To set up the static-dynamic framework, we partition the surrogate dataset $\mathcal{D}_\text{surrogate}$ into two subsets: $\mathcal{D}_\text{static}$ for the static surrogate stage and $\mathcal{D}_\text{dynamic}$ for the dynamic surrogate stage.

\textbf{(1) Static surrogate stage} In this initial stage, we train the prompts on $\mathcal{D}_\text{static} \cup \mathcal{D}_m$ to capture the label mapping functionality between the source and target datasets. During this phase, the prompts are optimized to instruct the model to correctly classify clean input images. Consequently, we obtain a pool of benign prompts for clean data. Denoting the prompt pool as $\mathbf{P}=\left\{\boldsymbol{p}_{1}, \boldsymbol{p}_{2}, \cdots, \boldsymbol{p}_{n_p}\right\}$ and $\mathbf{K}=\left\{\boldsymbol{k}_{1}, \boldsymbol{k}_{2}, \cdots, \boldsymbol{k}_{n_p}\right\}$ as the corresponding prompt keys, where $n_p$ is the prompt pool size, the objective for this optimization step follows \cite{wang2022learning} and is given by:

\begin{equation}
\label{equation:prompt_optimization}
\begin{aligned}
\min _{\mathbf{P}, \mathbf{K}, \phi} \mathcal{L}\left(\phi\left(f\left(\boldsymbol{x}; \boldsymbol{P}\right)\right), y\right)+\lambda \sum_{\mathbf{K}_{\boldsymbol{x}}} \gamma\left(q(\boldsymbol{x}), \boldsymbol{k}_{i}\right). \\
\end{aligned}
\end{equation}
Here, $\mathbf{K}_{\boldsymbol{x}}$ denotes a subset of the top-$K$ keys specifically selected for each sample $\boldsymbol{x}$. $\gamma$ is the function that assesses the similarity between the query feature $q(\boldsymbol{x})$ and prompt key. The scalar $\lambda$ weights the loss. The first term is the softmax cross-entropy loss, while the second term acts as a regularizer to encourage selected keys to be closer to the corresponding query features.

\textbf{(2) Trigger optimization stage} During this stage, the adversary optimizes the trigger $\boldsymbol{\delta}$ to induce misclassification of the triggered inputs into the target class. Specifically, the trigger loss function can be expressed as follows:

\begin{equation}
\label{equation:trigger_optimization}
\begin{aligned}
 \min _{\boldsymbol{\delta}} \sum_{(\boldsymbol{x}, c_m) \in \mathcal{D}_m}\left[\mathcal{L}\left(\phi\left(f\left(\boldsymbol{x} + \boldsymbol{\delta}; \boldsymbol{P}\right)\right), c_m\right)\right].
\end{aligned}
\end{equation}


\begin{figure}[!t]
  \centering
  \includegraphics[width=0.8\textwidth]{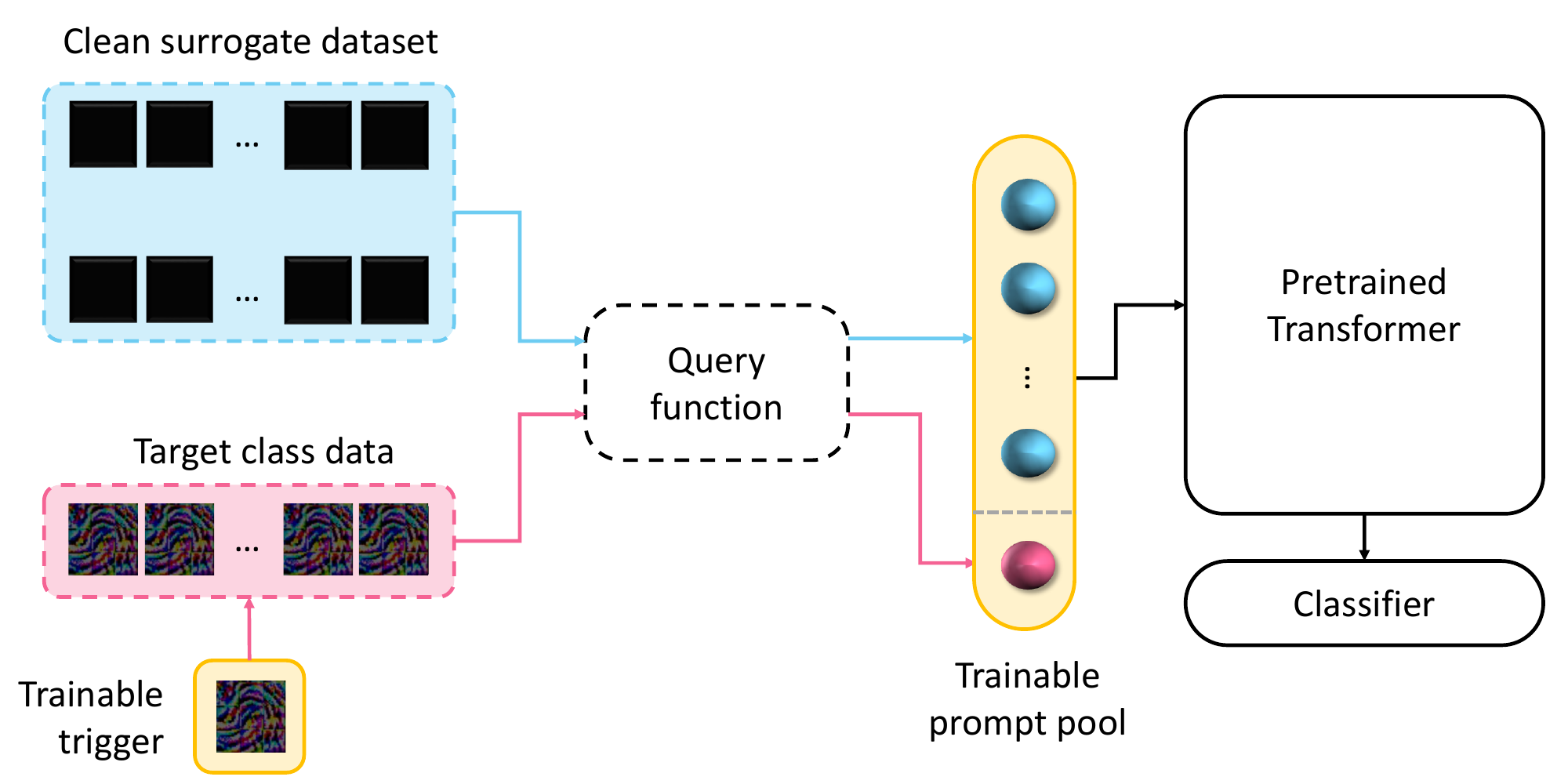}
  \caption{\textbf{AOP framework}. The backdoor trigger and prompt pool are updated using the static-dynamic strategy. Clean and poisoned data are mapped to corresponding prompts, guiding the pretrained model to behave normally on clean inputs while misclassifying triggered inputs according to the adversary's target.}
  \label{figure: training}
\end{figure}

\textbf{(3) Transition stage} This stage is designed to align the surrogate learner with the behaviour of the victim learner when being updated with new incoming tasks. Specifically, we continuously train the prompts from Stage \textbf{(1)} with the same objective as outlined in equation~\eqref{equation:prompt_optimization}, but using $\mathcal{D}_\text{dynamic}$. In essence, the goal of this stage is to statically prepare the surrogate learner for the subsequent dynamic stage.

\textbf{(4) Dynamic surrogate stage} In this stage, we aim to acquaint the backdoor trigger with the continuously updated prompts in continual learning. This dynamic stage entails fine-tuning the prompt components for one epoch, as in Stage \textbf{(3)}, following several iterations of optimization of the trigger with  equation~\eqref{equation:trigger_optimization}. This iterative process is repeated for multiple rounds to enhance the resilience of the backdoor trigger against the continual learning process. 

After optimizing the trigger through the aforementioned four stages, the optimized trigger $\boldsymbol{\delta}^{*}$ is used to poison a small portion of $\mathcal{D}_m$, which is then released to the victim learner. 
Summarization of AOP is in Figure \ref{figure: training} and further details are in the supplementary material.

\subsection{Towards an Authentic Backdoor Trigger}
\label{method:bce}

\textbf{Are we truly optimizing a backdoor trigger?} While our static-dynamic framework can generate a robust trigger that withstands intense incremental learning processes, it can unintentionally transform into adversarial noise. To further explore this phenomenon, we analyze the output scores in Figure \ref{fig:cheat_average_logits}. The visualization reveals that even when processed by a clean model unaffected by backdoor attacks, the poisoned samples are consistently misclassified towards the target class with dominant scores. This observation prompts a reconsideration of the backdoor trigger optimization process. We discovered that the overconfident score bias towards the target class is primarily induced by the commonly used softmax with cross-entropy loss function. Softmax introduces competition between classes, and the subsequent cross-entropy loss tends to elevate the scores of the target class significantly above the others. This pronounced bias compels the trigger to act like adversarial noise.

\textbf{Sigmoid with binary cross entropy loss.} To reduce biases, we mitigate the competition between the target class and other classes caused by the relative scoring of softmax by employing a sigmoid function after the logits to compute output scores. This approach shifts the optimization focus towards independently increasing the scores of target classes rather than suppressing others. Subsequently, we utilize binary cross-entropy loss to enable independent optimization processes. Following \cite{NEURIPS2021_5a9542c7}, the gradient of the loss at score $(s_{j})$ for class $j$ is computed as $ \frac{\partial \mathcal{L}_{\mathrm{BCE}}(\boldsymbol{\theta})}{\partial s_{j}} = \sigma(s_{j}) - \mathbb{I}\{j=\tilde{y}\} $, thereby constraining the score of the target class to a certain level regardless of the scores of other classes. As a result, during inference with a non-backdoored clean model, the output scores are more balanced between classes, as shown in Figure \ref{fig:bce_average_logits}. This balance prevents the problem of generating adversarial noise when optimizing the backdoor trigger.

\section{Experiments}
In this section, we first describe the experimental setup, then present the results in five key aspects: the overall backdooring ability of AOP, its performance with different surrogate datasets, the robustness of AOP with varying attack times, the efficacy of adopting BCE in preventing the generation of adversarial perturbations and its effectiveness compared to baselines. Further discussions on performance, visualizations, baselines, efficacy against defenses, and sensitivity to poisoning rates are deferred to the supplementary material.

\label{section:experiments}
\subsection{Experimental Setup}

\textbf{Victim incremental learners.} We evaluate our continual backdoor framework against 6 prompt-based continual learning methods: L2P \cite{wang2022learning}, DualPrompt \cite{wang2022dualprompt}, HiDe-Prompt \cite{wang2023hide}, CODA-Prompt \cite{smith2023codaprompt}, and two variants of PGP \cite{qiao2024PGP}, namely L2P-PGP and DualPrompt-PGP. We follow the original settings and implementations of each method. All learners utilize the ViT-B/16 backbone \cite{dosovitskiy2021an}, pre-trained on ImageNet-1K \cite{russakovsky2015imagenet}, except for HiDe-Prompt, which is pre-trained on iBOT-1K \cite{zhou2022image}. Detailed experimental information is in the supplementary material.

\begin{table*}[t!]
  \caption{Backdoor performance against L2P, DualPrompt, and PGP on 5-Split-CUB200. The attacker is the supplier for a random class in task $1$. The dynamic stage takes place over 5 rounds. Results are reported when using TinyImageNet and CIFAR100 as surrogate datasets. For ACC, we additionally report the change in clean accuracy compared to clean-training learners. For ASR, we provide a comparison with the baseline \cite{zeng2022narcissus} (without dynamic optimization and not using BCE).} 
  \label{table:cub200}
  \centering
  \begin{tabular}{lcccc}
    \toprule
    Surrogate dataset $\rightarrow$
    & \multicolumn{2}{c}{TinyImageNet}            & \multicolumn{2}{c}{CIFAR100}                   \\
    \cmidrule(r){2-3} \cmidrule(r){4-5} 
     & \multicolumn{1}{c}{ASR} & \multicolumn{1}{c}{ACC} & \multicolumn{1}{c}{ASR} & \multicolumn{1}{c}{ACC}  \\ 
    \midrule
L2P & $99.96 \pm 0.02$& $74.71 \pm 0.58$ & $99.99 \pm 0.02$ & $74.44 \pm 0.54$  \\
& $(\uparrow 86.44)$ & $(\downarrow 0.17)$& $(\uparrow 64.91)$ & $(\downarrow 0.44)$ \\
DualPrompt  & $99.93 \pm 0.02$ & $82.62 \pm 0.66$ & $99.95 \pm 0.05$ & $82.71 \pm 0.55$\\
& $(\uparrow 57.08)$ & $(\uparrow 0.10)$ & $(\uparrow 42.36)$ & $(\uparrow 0.19)$ \\
L2P-PGP &$99.97 \pm 0.01$&$74.97 \pm 0.83$& $100.00 \pm 0.00$ &$75.70 \pm 0.50$\\
& $(\uparrow 89.73)$ &$(\downarrow 0.48)$ & $(\uparrow 68.82)$ & $(\uparrow 0.25)$ \\
DualPrompt-PGP & $99.93 \pm 0.02$ & $82.45 \pm 0.29$ &$99.99 \pm 0.01$ &$82.84 \pm 0.12$ \\
& $(\uparrow 56.70)$ & $(\downarrow 0.31)$ & $(\uparrow 44.83)$ &  $(\uparrow 0.08)$ \\
    \bottomrule
  \end{tabular}
\end{table*}

\begin{table*}[t!]
  \caption{Backdoor performance across different prompt-based continual learning methods on three variants of Split-ImageNet-R. The adversary's target class is chosen randomly from the classes in task 1. The dynamic stage is iterated for 10 rounds. The surrogate dataset used is TinyImageNet. We also report the change in ACC compared to non-attacked learners.}
  \label{table:imagenetr}
  \centering
  \begin{tabular}{lcccccc}
    \toprule
    & \multicolumn{2}{c}{5-Split-ImageNet-R}            & \multicolumn{2}{c}{10-Split-ImageNet-R}           & \multicolumn{2}{c}{20-Split-ImageNet-R}           \\
    \cmidrule(r){2-3} \cmidrule(r){4-5} \cmidrule(r){6-7}
     & \multicolumn{1}{c}{ASR} & \multicolumn{1}{c}{ACC} & \multicolumn{1}{c}{ASR} & \multicolumn{1}{c}{ACC} & \multicolumn{1}{c}{ASR} & \multicolumn{1}{c}{ACC} \\ 
    \midrule
L2P & $99.76 \pm 0.10$ & $64.27 \pm 0.65$ & $99.56 \pm 0.22 $  & $ 62.43 \pm 0.58 $ & $98.24 \pm 0.21$ & $60.51 \pm 1.17$ \\
& & $(\downarrow 0.77) $ & & $(\downarrow 0.12)$ & & $(\downarrow 0.83)$ \\
DualPrompt & $99.57 \pm 0.25$ & $70.69 \pm 0.56$ & $99.26 \pm 0.39 $ & $69.17 \pm 0.27$ & $96.17 \pm 0.89$ & $66.04 \pm 0.43$  \\
& & $(\downarrow 0.62)$ & & $(\downarrow 0.85)$ & & $(\downarrow 0.21)$ \\
CODA-Prompt & $98.16 \pm 1.01$ & $74.15 \pm 0.11$ & $96.55 \pm 1.29$ & $72.86 \pm 0.11$ & $71.27 \pm 2.86$ &   $70.86 \pm 0.94$                      \\
& & $(\downarrow 1.04)$ & & $(\downarrow 0.02)$ & & $(\downarrow 0.04)$ \\
HiDe-Prompt & $98.65 \pm 0.90$ & $74.89 \pm 0.60$ & $94.66 \pm 0.93$ &  $71.99 \pm 0.37$ & $93.79 \pm 0.66$ &  $70.93 \pm 0.86$                       \\
& & $(\downarrow 0.32)$ & & $(\downarrow 0.46)$ & & $(\downarrow 0.09)$ \\
L2P-PGP     & $99.33 \pm 0.05$ & $64.38 \pm 0.57$ & $99.36 \pm 0.15$ & $61.73 \pm 0.38 $ & $98.84 \pm 0.16$  & $60.74 \pm 1.17$ \\
& & $(\uparrow 0.10)$ & & $(\uparrow 0.33)$ & & $(\downarrow 0.15)$ \\
DualPrompt-PGP & $99.83 \pm 0.27$ & $70.80 \pm 0.08$ & $99.17 \pm 0.43$ & $69.24 \pm 0.41$ & $97.01 \pm 0.75$ & $66.32 \pm 1.04$ \\
& & $(\downarrow 0.08)$ & & $(\downarrow 0.18)$ & & $(\downarrow 0.76)$ \\
    \bottomrule
  \end{tabular}
\end{table*}

\textbf{Datasets.} For the victim's training dataset, we follow existing prompt-based continual learning methods \cite{wang2023hide, qiao2024PGP} and use three variants of ImageNet-R \cite{Hendrycks2020TheMF}: 5-Split, 10-Split, and 20-Split ImageNet-R. These variants divide the 200 classes of the original dataset into 5, 10, and 20 tasks, respectively. Additionally, we conduct experiments on the 5-Split-CUB200 dataset, which partitions the original CUB200 \cite{WahCUB_200_2011} dataset into 5 tasks, each containing 40 classes. For the attacker's surrogate dataset, we primarily use TinyImageNet \cite{Le2015TinyIV} for all experiments and CIFAR100 \cite{Krizhevsky2009LearningML} in specific settings.

\textbf{Backdoor setting.} Following the guidelines of \cite{zeng2022narcissus}, we set the maximum poison ratio to 25 images, corresponding to $0.1\%$ of ImageNet-R and $0.5\%$ of CUB200. Additionally, we set the upper bound of the \(\ell_\infty\)-norm of triggers to \(\frac{16}{255}\), in line with standard practices in the literature \cite{turner2019labelconsistent, saha2019hidden}. During inference, the trigger is amplified by a factor of $3$ \cite{turner2019labelconsistent, zeng2022narcissus}.

\textbf{Metrics.} The evaluation of our framework utilizes two key metrics: (1) average accuracy (ACC) and (2) attack success rate (ASR). ACC assesses the accuracy of the backdoored model on benign test samples, whereas ASR measures the proportion of attacked samples that the compromised model predicts as the target label, reflecting the backdoor attack's effectiveness. In the context of continual learning, ACC and ASR at a given time $t$ are averaged across the corresponding metrics for all data from task $1$ to task $t$. All results are averaged over 3 runs for fair comparisons.

\subsection{Effectiveness of AOP} 
\label{section:main_results}

We report the ASR and ACC when performing backdoor attacks against various incremental learners in Table \ref{table:cub200} and Table \ref{table:imagenetr}. As observed from the tables, our framework consistently achieves high ASR with negligible effect on the ACC of clean samples. This is due to the inherent characteristics of continual learning, which enable the learner to perform well across different tasks, making it vulnerable to backdoor attacks. By considering backdooring in continual learning as an additional "backdoor task," the plasticity of continual learning allows the ASR, or performance on the backdoor task, to be high without degrading the ACC on clean tasks.

It is worth noting that ASR still suffers from the catastrophic forgetting phenomenon of continual learning for long sequence tasks. Specifically, in Table \ref{table:imagenetr}, the 20-Split-ImageNet-R performs worse than the 5-split and 10-split versions across all experiments. This indicates that the more tasks and the longer the incremental learning process, the higher the chance for a decrease in ASR. However, the ACC also suffers from this phenomenon, as it is a major issue in continual learning. 

While prompt-based methods share a common core of utilizing prompt pools and selecting relevant prompts for each task or class, each exhibits distinct characteristics. AOP observes a significantly lower ASR when backdooring CODA-Prompt. This is because CODA-Prompt utilizes all prompts in the prompt pool through its weighted mechanism instead of selecting only the top-K relevant prompts. Consequently, even with triggered samples, clean prompts still exert some influence, leading to degradation in ASR.

\begin{table*}[t!]
  \caption{Backdoor performance when the target class belongs to different tasks $T$. The results are reported when the victim's training dataset is 10-Split-ImageNet-R, and the attacker's surrogate dataset is TinyImageNet.}
  \label{table:tasks}
  \centering
  \begin{tabular}{lcccccc}
    \toprule
    & \multicolumn{2}{c}{$T=1$}            & \multicolumn{2}{c}{$T=4$}           & \multicolumn{2}{c}{$T=10$}           \\
    \cmidrule(r){2-3} \cmidrule(r){4-5} \cmidrule(r){6-7}
     & \multicolumn{1}{c}{ASR} & \multicolumn{1}{c}{ACC} & \multicolumn{1}{c}{ASR} & \multicolumn{1}{c}{ACC} & \multicolumn{1}{c}{ASR} & \multicolumn{1}{c}{ACC} \\ 
    \midrule
L2P & $99.56 \pm 0.22 $  & $ 62.43 \pm 0.58 $ & $99.61 \pm 0.19$&  $62.09 \pm 0.06$ &   $99.89 \pm 0.05$& $62.27 \pm 0.26$ \\
L2P-PGP & $99.36 \pm 0.15$ & $62.73 \pm 0.38 $ & $99.77 \pm 0.08$ &  $62.88 \pm 0.73$& $99.85 \pm 0.35$& $62.32 \pm 0.82$ \\
    \bottomrule
  \end{tabular}
\end{table*}

\begin{table*}[t!]
  \caption{ASR of clean, non-attacked learners on triggered samples. Results are compared between triggers optimized with CE softmax and BCE sigmoid loss.}
  \label{table:cheating}
  \centering
  \begin{tabular}{llcccc}
    \toprule
    & & \multicolumn{1}{c}{L2P} & \multicolumn{3}{c}{DualPrompt}           \\
     \cmidrule(r){3-3}  \cmidrule(r){4-6} 
     &  & \multicolumn{1}{c}{10-Split-} & \multicolumn{1}{c}{5-Split-} & \multicolumn{1}{c}{10-Split-}
     & \multicolumn{1}{c}{20-Split-} \\ 
     &  & \multicolumn{1}{c}{ImageNet-R} & \multicolumn{1}{c}{ImageNet-R} & \multicolumn{1}{c}{ImageNet-R}
     & \multicolumn{1}{c}{ImageNet-R} \\ 
     
    \midrule
AOP with CE & Top-1 ASR &  $74.18$ & $34.18$ & $42.85$ & $96.93$ \\
 & Top-5 ASR &  $96.89$ & $92.78$ & $97.01$ & $99.63$ \\
\cmidrule(r){2-6}
AOP with BCE & Top-1 ASR & $0.00$ & $0.00$ & $0.00$ & $0.00$ \\
 & Top-5 ASR & $0.00$ & $0.72$ & $0.12$ & $2.68$ \\
    \bottomrule
  \end{tabular}
\end{table*}

\textbf{Different surrogate datasets.} Another factor that makes prompt-based continual learning vulnerable is the utilization of prompting. As shown in Figures \ref{fig:clean_correl} and \ref{fig:ba_correl}, AOP's triggered samples consistently have the highest similarity with prompt ID 2, which, in contrast, shows the smallest similarity with benign samples. Thus, as discussed in Section \ref{method:transferability}, prompting allows for actual data differences when choosing surrogate datasets. We report the backdoor performance using TinyImageNet and CIFAR100 as surrogate datasets in Table \ref{table:cub200}. The experiments show consistently high ASR results for both surrogate data choices, confirming the transferability of our continual backdoor framework.

\textbf{Different attack times.} We report the ASR in Table \ref{table:tasks}, considering scenarios where the target class belongs to different tasks that arrive at different times. We observe slight increases in ASR when the attack class is part of later tasks, as it experiences less forgetting. Nonetheless, our method AOP consistently maintains a high ASR, exceeding $99\%$ at all three reported attack times. This convincingly demonstrates that the backdoor knowledge can be effectively transferred to both previously learned and incoming future classes.

\begin{figure}[!t]
     \centering
     \begin{subfigure}[b]{0.22\textwidth}
         \centering
         \includegraphics[width=\textwidth]{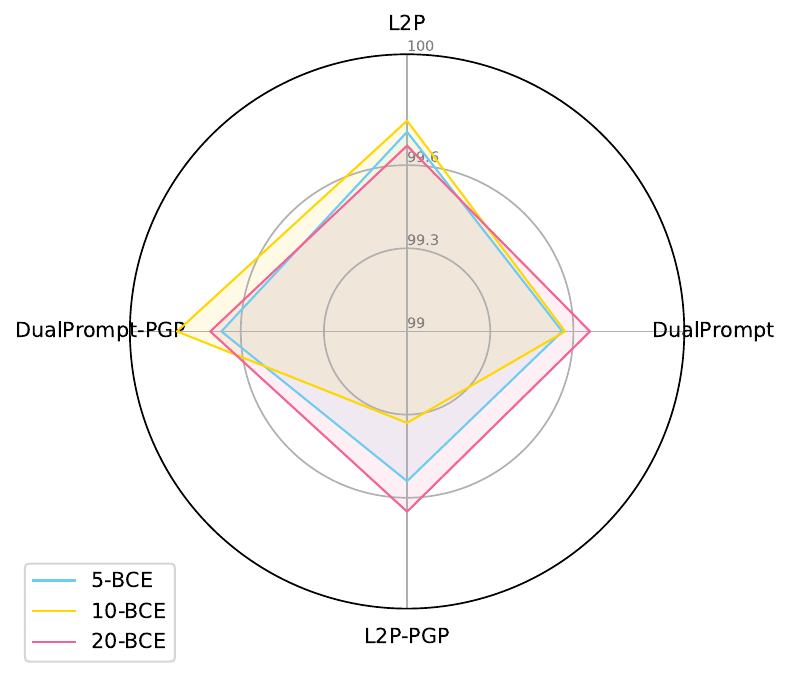}
         \caption{5-Split-ImageNet-R}
         \label{rounds:5tasks}
     \end{subfigure}
     \begin{subfigure}[b]{0.22\textwidth}
         \centering
         \includegraphics[width=\textwidth]{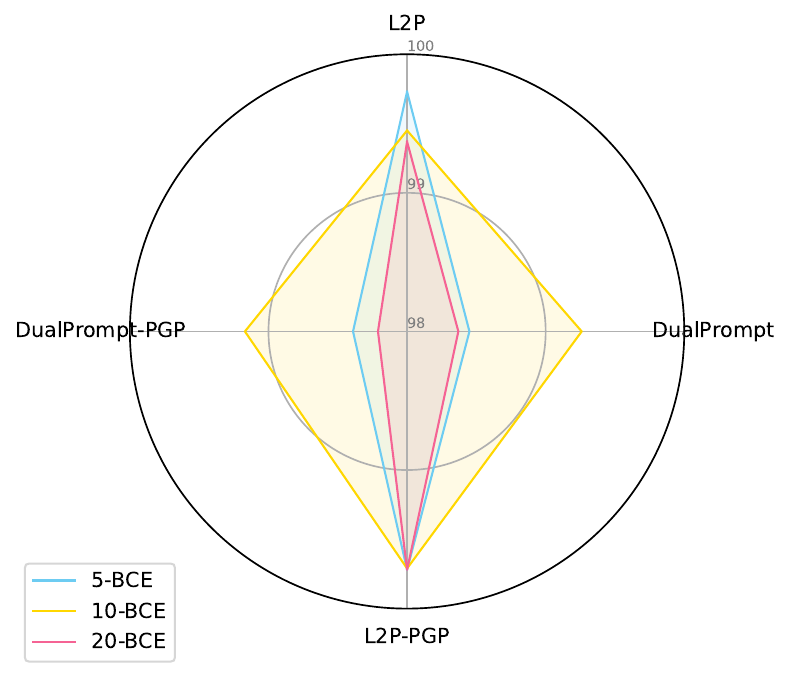}
         \caption{10-Split-ImageNet-R}
          \label{rounds:10tasks}
     \end{subfigure}
     \begin{subfigure}[b]{0.22\textwidth}
         \centering
         \includegraphics[width=\textwidth]{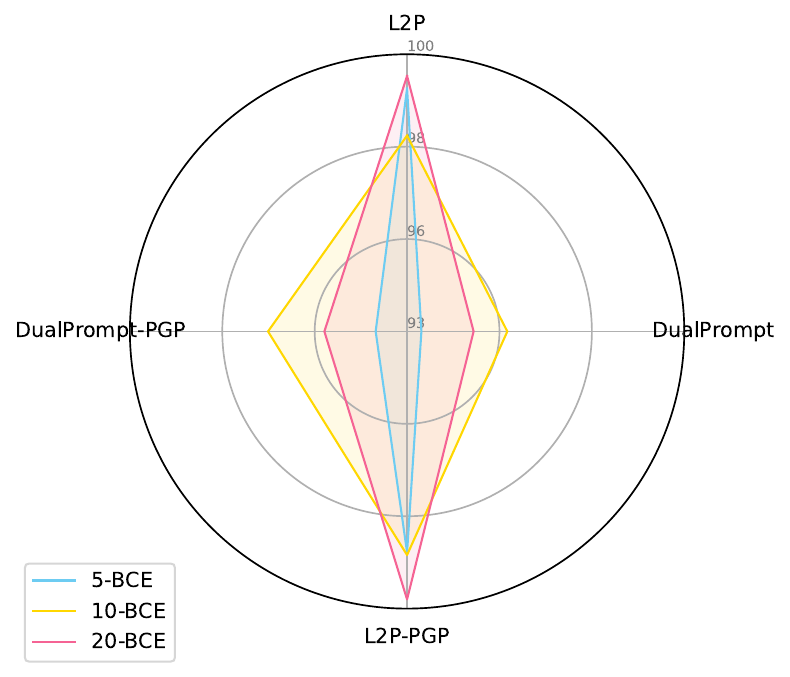}
         \caption{20-Split-ImageNet-R}
          \label{rounds:20tasks}
     \end{subfigure}
     \caption{ASR when varying number of dynamic rounds.}
      \label{fig:rounds}
\end{figure}

\textbf{Different dynamic rounds.} We illustrate the attack performance across varying numbers of dynamic rounds in Figure \ref{fig:rounds}. As discussed above, the ASR decreases when tested on the 20-Split-ImageNet. 
We observe that increasing the number of dynamic rounds does not consistently lead to higher performance. However, from a positive perspective, since the adversary is unaware of the total tasks, adjusting the number of dynamic rounds should minimally impact ASR. We emphasize that in long sequence tasks, both ASR and ACC degrade due to forgetting.

\textbf{Enhancing backdoor authenticity via sigmoid BCE.} As shown in Table \ref{table:cheating}, triggers optimized with softmax CE retain considerable scores even when tested on non-backdoored models. This suggests that CE optimization might lead to the generation of adversarial perturbations. Conversely, when optimized using sigmoid BCE, the ASR on clean models remains consistently low. This confirms that adopting BCE can enhance the authenticity of backdoor triggers and avoid generating adversarial noise.

\begin{figure}[!t]
  \centering
  \includegraphics[width=0.48\textwidth]{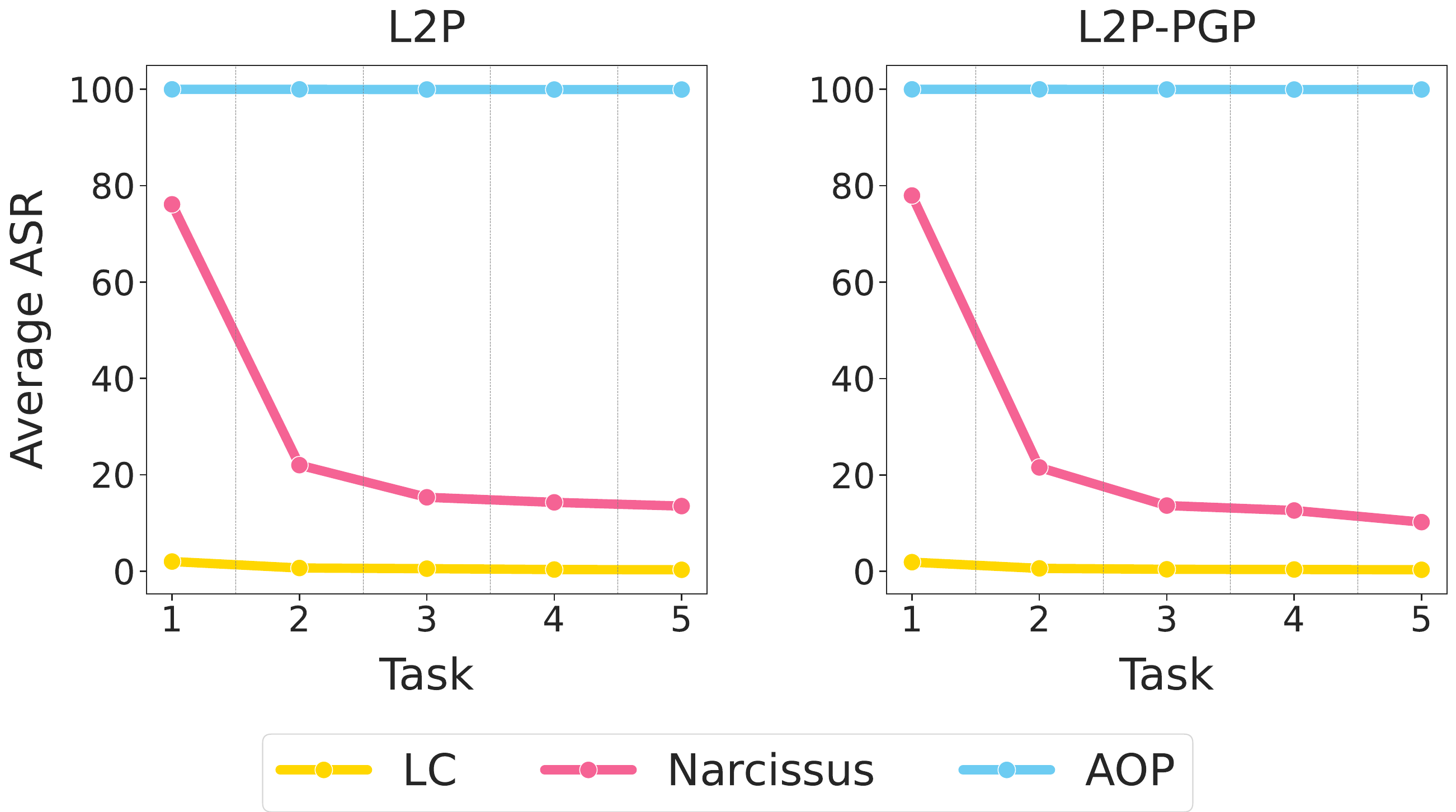}
  \caption{Average ASR after each task during the attack on L2P and L2P-PGP on CUB200. The figure compares the performance of LC (Label Consistent) \cite{turner2019labelconsistent}, Narcissus \cite{zeng2022narcissus}, and AOP (ours).}
  \label{figure: asr}
\end{figure}

\textbf{Comparisons to baselines.} We compare the performance of AOP in executing backdoor attacks on prompt-based continual learners against the black-box and clean-label backdoor frameworks, including Narcissus \cite{zeng2022narcissus} and Label Consistent (LC) \cite{turner2019labelconsistent}. As illustrated in Figure \ref{figure: asr}, during the first task, both AOP and Narcissus significantly outperform LC, as LC is not designed for data-constrained scenarios. However, as the incremental learning process progresses, Narcissus suffers from forgetting, leading to a sharp decline in its average ASR. In contrast, AOP maintains its backdoor effectiveness throughout all tasks, thanks to its dynamic training strategy.

\section{Conclusion}
\label{section:conclusion}
This paper explores the vulnerability of prompt-based continual learning methods and their susceptibility to backdoor attacks. We emphasize three critical properties that a backdoor continual framework should possess: transferability to unknown data from other classes, resilience against incremental learning procedures, and the authenticity of the backdoor trigger. Building upon these considerations, we propose a novel continual backdoor framework. We leverage the label mapping functionality of prompting to promote transferability, incorporate a static-dynamic optimization approach to enhance resilience, and employ BCE sigmoid loss to mitigate the adversarial noise problem. Extensive experiments confirm the effectiveness of our backdoor framework against various prompt-based continual learners. 

Nonetheless, we acknowledge some limitations in our work. Firstly, competition between the target classes and the remaining classes remains necessary to some extent. Relying solely on BCE to eliminate relative scoring might hurt the performance. Secondly, certain defenses we employed to assess our approach may not be optimal for continual learning scenarios. Thus, regarding future directions, there is potential in exploring other threat models and defenses for backdooring continual learning and extending backdoor attacks to other continual learning approaches.
\bibliography{references}
\bibliographystyle{abbrv}



\appendix
\newpage

In this supplementary material, we first review related work on continual learning, prompt-based continual learning, and backdoor attacks in Appendix \ref{section:related_work}. Next, we summarize our AOP in Appendix \ref{section:aob_apx}. Implementation details, additional experiments, and visualizations are provided in Appendices \ref{section:imlementation_details} and \ref{section:additional_experiments}, respectively. Finally, we discuss broader impacts in Appendix \ref{appendix: broader impacts}.

\section{Related Work}
\label{section:related_work}

\paragraph{Continual learning} Adapting to new knowledge is an innate human capability, yet it poses significant challenges for machine learning models. Continual learning emerges as one approach to bridge this gap between models and humans, which encourages models to continuously acquire new knowledge from new data while retaining previously learned ones. The regularization/prior approach \cite{Kirkpatrick_2017, nguyen2018variational, farquhar2019unifying, loo2020generalized, jung2021continual, ahn2019uncertaintybased, yin2020sola} effectively preserves learned knowledge by controlling the learning of the model's parameters through a regularization term in the objective function. Architecture-based approaches \cite{li2019learn, mallya2018packnet, serrà2018overcoming, wang2020learnpruneshare, yan2021der, abati2020conditional, Liu_2021} extend the model's plasticity by expanding its network to accommodate new knowledge. Rehearsal-based approaches \cite{buzzega2020dark, cha2021co2l, pham2023continual, chaudhry2019tiny} rely on a memory buffer to retain past knowledge. Continual learning primarily focuses on the class-incremental learning (CIL) setting, which is the most challenging and representative setting since the task boundaries are not available during inference. While rehearsal-based approaches achieve state-of-the-art performance \cite{buzzega2020dark} in CIL, they violate data privacy requirements as they necessitate the storage of past data.

\paragraph{Prompt-based continual learning} 
With few learnable parameters and not relying on memory buffers, prompt-based continual learning methods achieve state-of-the-art performance. These methods are especially suitable for scenarios where data privacy is crucial. Specifically, prompt-based approaches leverage the power of pre-trained models, learning only a small number of prompts to guide the model's performance across different tasks or classes. L2P \cite{wang2022learning} is the first work to explore prompting in continual learning. It constructs a prompt pool and selects appropriate prompts for each input. Building on L2P, DualPrompt \cite{wang2022dualprompt} employs prefix-tuning and constructs two types of prompts: task-sharing and task-specific. CODA-Prompt \cite{smith2023codaprompt} enhances prompt selection with an adaptive attention mechanism. HiDe-Prompt \cite{wang2023hide} examines the influence of various pretraining paradigms and decomposes the objective into hierarchical components. PGP \cite{qiao2024PGP} uses prompt gradient projection to promote updates in orthogonal directions, effectively preventing forgetting.

\paragraph{Backdoor attack}

A backdoor attack aims to cause a model to misbehave according to an adversary's target when the input data contains a specific backdoor trigger, while still performing normally on clean input data. Backdoor attacks have been explored in different settings and under various threat models, which identify the attacker's accessibility. In a black-box setting \cite{saha2019hidden, souri2021sleeper, li2022untargeted, sun2024backdoor}, the attacker has no control over the training process and only has access to the dataset, which they then poison and release to the victim. Another line of work \cite{saha2019hidden, souri2021sleeper, zeng2022narcissus, turner2019labelconsistent} assumes that the attacker cannot flip the labels of the dataset (clean-label). Recently, attackers' control has been limited to data-constrained scenarios where they only have access to a small proportion of data. For example, \cite{zeng2022narcissus} employs a surrogate clean model to optimize a clean-label backdoor trigger, while \cite{li2024efficient} leverages the zero-shot capabilities of the CLIP model to suppress clean features and augment the poisoning features. Additionally, \cite{Huynh_Nguyen_Pham_Tran_2024} observes that even with carefully alternated training to train a surrogate poisoned model, the optimized backdoor trigger tends to become adversarial noise.

Previous works on backdoor attacks against continual learning have primarily focused on non-targeted attacks, aiming to degrade the model's performance in general. These studies typically explore task-incremental and domain-incremental settings using various approaches. For instance, \cite{9892774} describes a white-box attack where the attacker has control over the training model and seeks to force the neural network to forget previously learned knowledge. Other works, such as \cite{9206809, Umer2021AdversarialTF, Umer2022FalseMF}, focus on regularization-based and replay-based learners in domain-incremental and task-incremental learning scenarios, aiming to degrade the performance of the first task. Similarly, \cite{9892774} and \cite{pmlr-v202-kang23c} aim to undermine the performance of continual learners. In contrast, our work focuses on targeted backdoor attacks. We aim to manipulate the attacked learner to classify poisoned data from any task into a desired target class while maintaining high accuracy on clean data. Furthermore, our research emphasizes state-of-the-art prompt-based continual learning and tackles the most challenging setting in continual learning, which is class-incremental learning.


\section{AOP end-to-end pipeline}
\label{section:aob_apx}

In this Appendix, we provide an overview of the key algorithms utilized in AOP. Specifically, Algorithm \ref{alg:prompt_tuning} details the process for prompt tuning, Algorithm \ref{alg:trigger_gen} outlines the method for trigger optimization, and Algorithm \ref{alg:full} presents the comprehensive end-to-end pipeline of AOP.

\begin{figure*}
     \centering
     \begin{subfigure}[b]{0.30\textwidth}
         \centering
         \includegraphics[width=\textwidth]{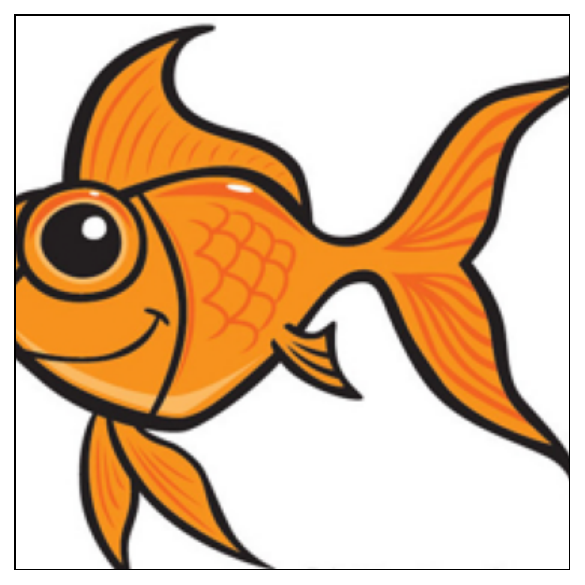}
         \caption{Clean image}
     \end{subfigure}
     \begin{subfigure}[b]{0.30\textwidth}
         \centering
         \includegraphics[width=\textwidth]{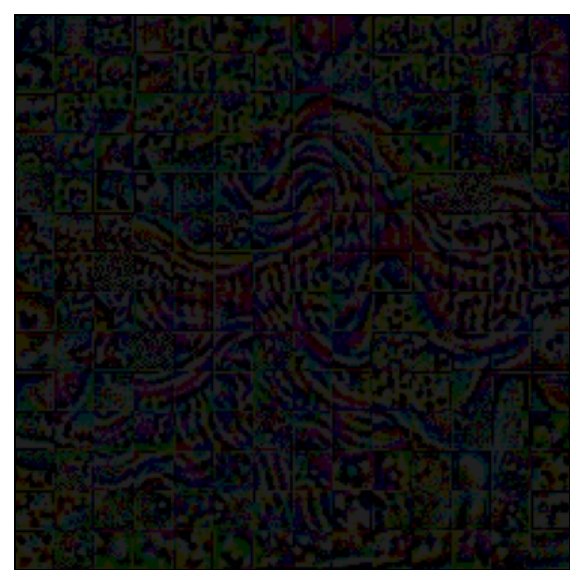}
         \caption{Backdoor trigger}
     \end{subfigure}
     \begin{subfigure}[b]{0.30\textwidth}
         \centering
         \includegraphics[width=\textwidth]{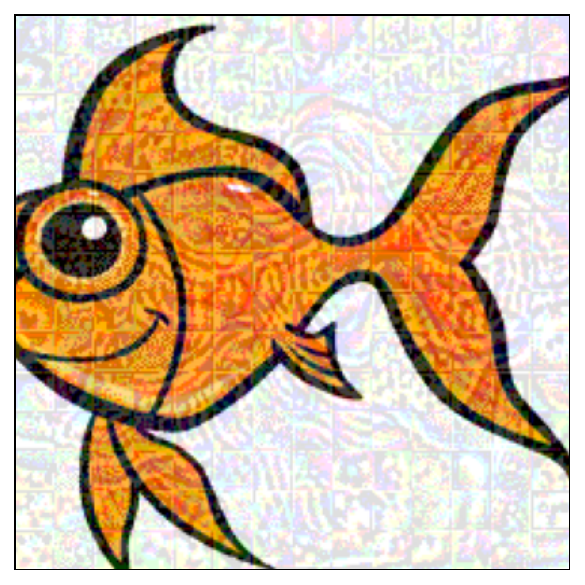}
         \caption{Poisoned image}
     \end{subfigure}
     \caption{Visualizations of the clean image, backdoor trigger, and poisoned image.}
      \label{fig:noise}
\end{figure*}

\begin{algorithm}[hbt!]
\caption{Prompt Tuning}\label{alg:prompt_tuning}
\KwInput{(1) Surrogate model $f$}
\myinput{(2) Dataset $\mathcal{D}$}
\myinput{(3) Prompt components $ \boldsymbol{P} = \left\{\left(\boldsymbol{k}_{1}, \boldsymbol{p}_{1}\right),\left(\boldsymbol{k}_{2}, \boldsymbol{p}_{2}\right), \cdots,\left(\boldsymbol{k}_{n_p}, \boldsymbol{p}_{n_p}\right)\right\}$}
\myinput{(4) Query function $q$}
\myinput{(5) Cosine similarity $\gamma$}
\myinput{(6) Top-K selected keys $\mathbf{K}_{\boldsymbol{x}}$}
\myinput{(7) Number of iterations for trigger generating $\mathcal{K}$}
\myinput{(8) Learning rate $\alpha > 0$}
\KwOutput{The optimized prompts $\boldsymbol{P}^{*}$}
\Comment{Initialization}
Initialize with input $\boldsymbol{P}$;

$k \leftarrow 0$\;
\While{$k < \mathcal{K}$}{
    \Comment{Update prompts}
    $\boldsymbol{P}^{k+1} \leftarrow \boldsymbol{P}^{k} - \alpha \sum_{(\boldsymbol{x},y) \in \mathcal{D}} \nabla_{\boldsymbol{P}} \mathcal{L} (f(\boldsymbol{x}; \boldsymbol{P}), y) - \lambda \sum_{\mathbf{K}_{\boldsymbol{x}}} \gamma\left(q(\boldsymbol{x}), \boldsymbol{k}_{i}\right)$
    }
\KwReturn{$\boldsymbol{P}^{*}$}
\end{algorithm}

\begin{algorithm}[hbt!]
\caption{Trigger Optimization}\label{alg:trigger_gen}
\KwInput{(1) Surrogate model $f$}
\myinput{(2) Target class data samples $\mathcal{D}_{m}=\left\{\left(\boldsymbol{x}, y\right) \mid y = c_m\right\}$}
\myinput{(3) Prompt components $ \boldsymbol{P}$}
\myinput{(4) Trigger $ \boldsymbol{\delta}$}
\myinput{(5) Criterion $ \boldsymbol{\zeta}$}
\myinput{(6) Allowable set of trigger patterns $\Delta$}
\myinput{(7) Number of iterations for prompt tuning $\mathcal{I}$}
\myinput{(8) Learning rate $\eta > 0$}
\KwOutput{The optimized adaptive trigger $\boldsymbol{\delta^{*}}$}
\Comment{Initialization}
$\boldsymbol{\delta}_{0} \leftarrow \boldsymbol{\delta}$\;
$i \leftarrow 0$\;
\While{$i < \mathcal{I}$}{
    \Comment{Update trigger}
    $\boldsymbol{\delta}_{i+1} \leftarrow \boldsymbol{\delta}_{i} - \eta \sum_{(x,c_m) \in \mathcal{D}_{m}} \nabla_{\boldsymbol{\delta}} \mathcal{L} (f(\boldsymbol{x} + \boldsymbol{\delta};\boldsymbol{P}), c_m)$;
    
    \Comment{Constraint trigger in $\ell_{p}$-norm ball}
    $\boldsymbol{\delta}_{i+1} \leftarrow Proj_{\Delta}(\boldsymbol{\delta}_{i+1})$
} 
\KwReturn{$\boldsymbol{\delta}^{*}$}
\end{algorithm}

\begin{algorithm}[hbt!]
\caption{AOP End-to-end Pipeline}\label{alg:full}
\KwInput{(1) Initial surrogate model $f$}
\myinput{(2) Prompt pool $\mathbf{P}$}
\myinput{(3) Target class $c_m$}
\myinput{(4) Target class data samples $\mathcal{D}_{m}$}
\myinput{(5) Surrogate $\mathcal{D}_{surrogate}$}
\myinput{(6) Number of iterations for full optimization $\mathcal{E}$}
\KwOutput{The optimized adaptive trigger $\boldsymbol{\delta}^{*}$}

\Comment{Partition the surrogate datasets into two subsets.}
$\mathcal{D}_\text{surrogate} = \mathcal{D}_\text{static} \cup \mathcal{D}_\text{dynamic}$

\Comment{Static surrogate stage.}
$\mathbf{P}$ $\leftarrow$ PromtTuning($f$, $\mathcal{D}_{static} \cup \mathcal{D}_m$,  $\mathbf{P}$)

$ $
\Comment{Static trigger optimization}
$Initialize \boldsymbol{\delta}$\;
$\boldsymbol{\delta} \leftarrow$ TriggerUpdate($f, \mathcal{D}_{m}, \mathbf{P}$, $\zeta$ = CE);

\Comment{Transition stage}
$\mathbf{P}$ $\leftarrow$ PromtTuning($f$, $\mathcal{D}_{dynamic}$,  $\mathbf{P}, \boldsymbol{\delta}$)

\Comment{Dynamic stage}
\While{$e < \mathcal{E}$}{
    \Comment{Update trigger}
    $\boldsymbol{\delta}_{e+1} \leftarrow$ TriggerUpdate($f, \mathcal{D}_{m}, \mathbf{P}_{e}, \boldsymbol{\delta}_{e}$, $\zeta$ = BCE);

    \Comment{Update malicious prompt}
    $\boldsymbol{P}_{ e+1} \leftarrow$ PromptTuning($f, \mathcal{D}_{dynamic}, \mathbf{P}_{e}$);
    
} 
\KwReturn{$\boldsymbol{\delta}^{*}$}
\end{algorithm}

\section{Implementation Details}
\label{section:imlementation_details}


In this Section, we provide the implementation details of all experiments.

\paragraph{Victim prompt-based Learners} Our implementations of L2P, DualPrompt, L2P-PGP, and DualPrompt-PGP are based on the source code provided by \cite{qiao2024PGP}. The implementations of HiDe and CODA-Prompt are based on the original papers by \cite{wang2023hide} and \cite{smith2023codaprompt}, respectively. All experiments were conducted on NVIDIA V100 GPUs. For all victim learners, we utilize the Adam optimizer with $\beta_1 = 0.9$ and $\beta_2 = 0.999$.

For the L2P and L2P-PGP methods, we train the victim learner on the 5-Split-CUB200 dataset for 5 epochs per task, using a batch size of 16 and a prompt length of 5. When training on the 5/10/20-Split-ImageNet-R datasets, the number of epochs per task increases to 50, with a prompt length of 20. For DualPrompt and DualPrompt-PGP, training on the 5-Split-CUB200 dataset involves 10 epochs per task, with a prompt length of 5 and a batch size of 24. For the 5/10/20-Split-ImageNet-R datasets, these methods are trained for 50 epochs per task, with a prompt length of 20 and a batch size of 24. The HiDe-Prompt method employs 10 prompts, each with a length of 20, across all Split-ImageNet-R variants, training the main architecture for 50 epochs with a batch size of 24. Lastly, the CODA-Prompt method uses a configuration with 50 prompts, a pool size of 100, and a prompt length of 8.

The training times for the 6 incremental learners on the 5/10/20-SplitImageNet-R dataset range from 8 to 10 hours. For the Split-CUB200 dataset, the training times for L2P, L2P-PGP, DualPrompt, and DualPrompt-PGPP are 0.5 hours, 1 hour, 1.5 hours, and 2 hours, respectively.

\begin{figure*}[!t]
  \centering
  \includegraphics[width=\textwidth]{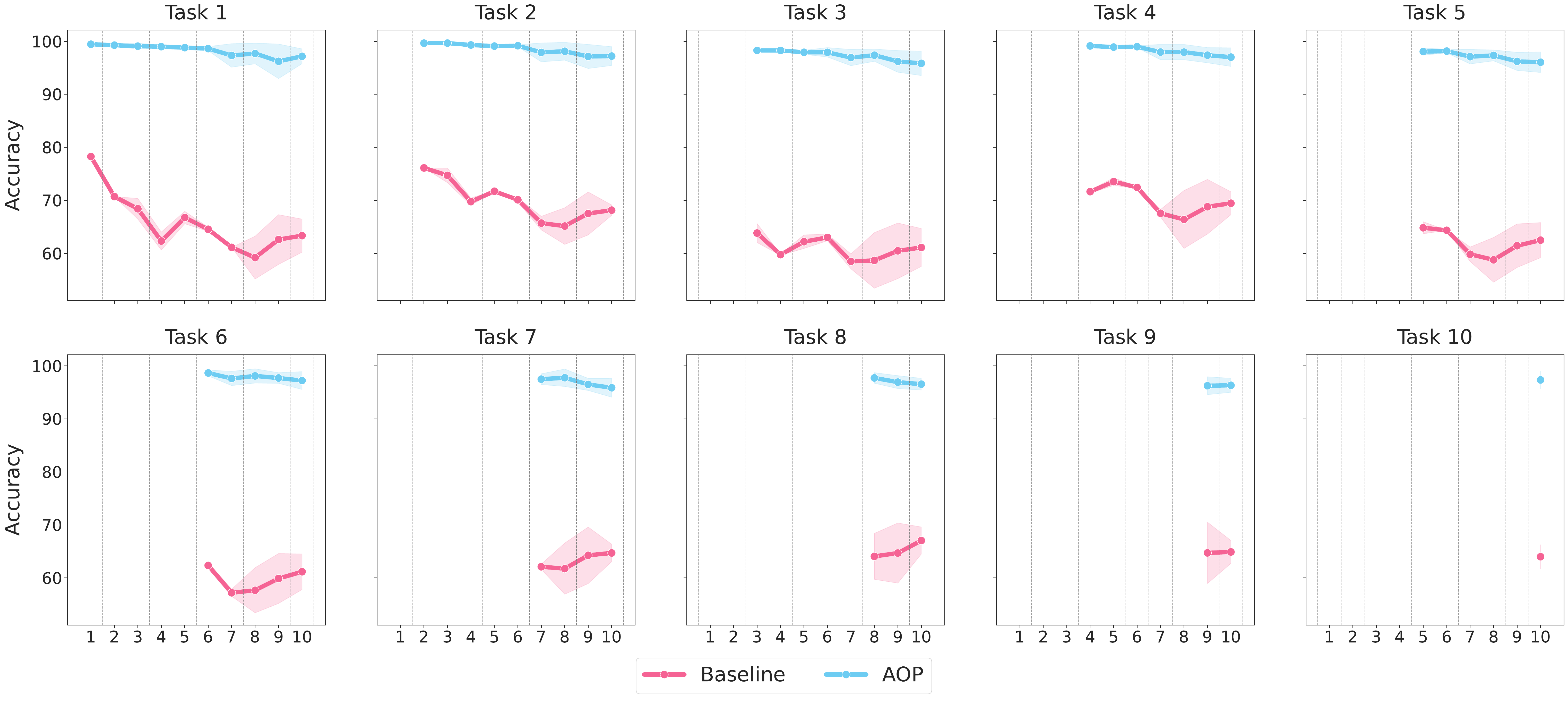}
  \caption{Comparison of ASR history for each task during the incremental learning process between AOP and Narcissus, using CODA-Prompt with 10-Split-ImageNet-R dataset for visualization.}
  \label{fig: asr_flow}
\end{figure*}

\begin{figure}[!t]
  \centering
  \includegraphics[width=0.4\textwidth]{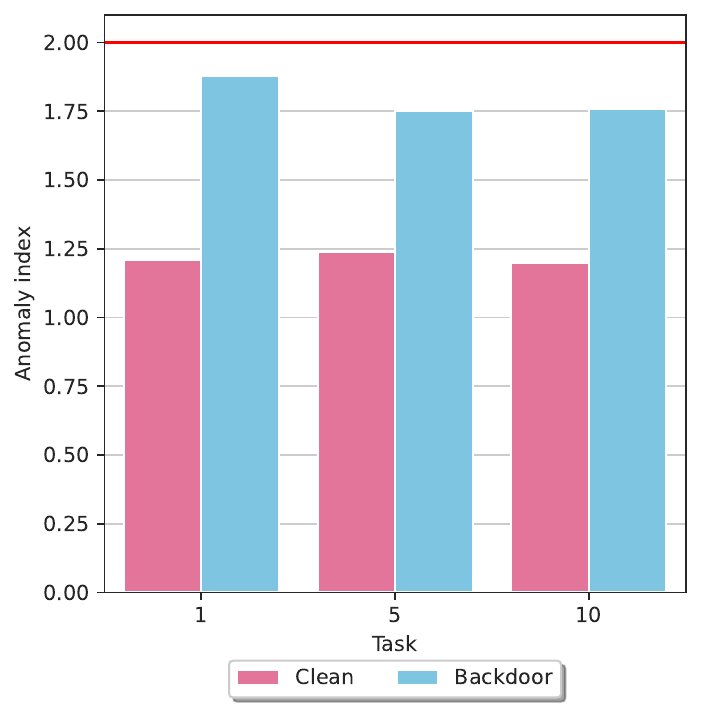}
  \caption{Evaluation of AOP against Neural Cleanse. Results are reported at three checkpoints: tasks 1, 5, and 10, when attacking L2P on Split-ImageNet-R.}
  \label{fig: neural_cleanse}
\end{figure}

\begin{figure*}[!t]
     \centering
     \begin{subfigure}[b]{0.19\textwidth}
         \centering
         \includegraphics[width=\textwidth]{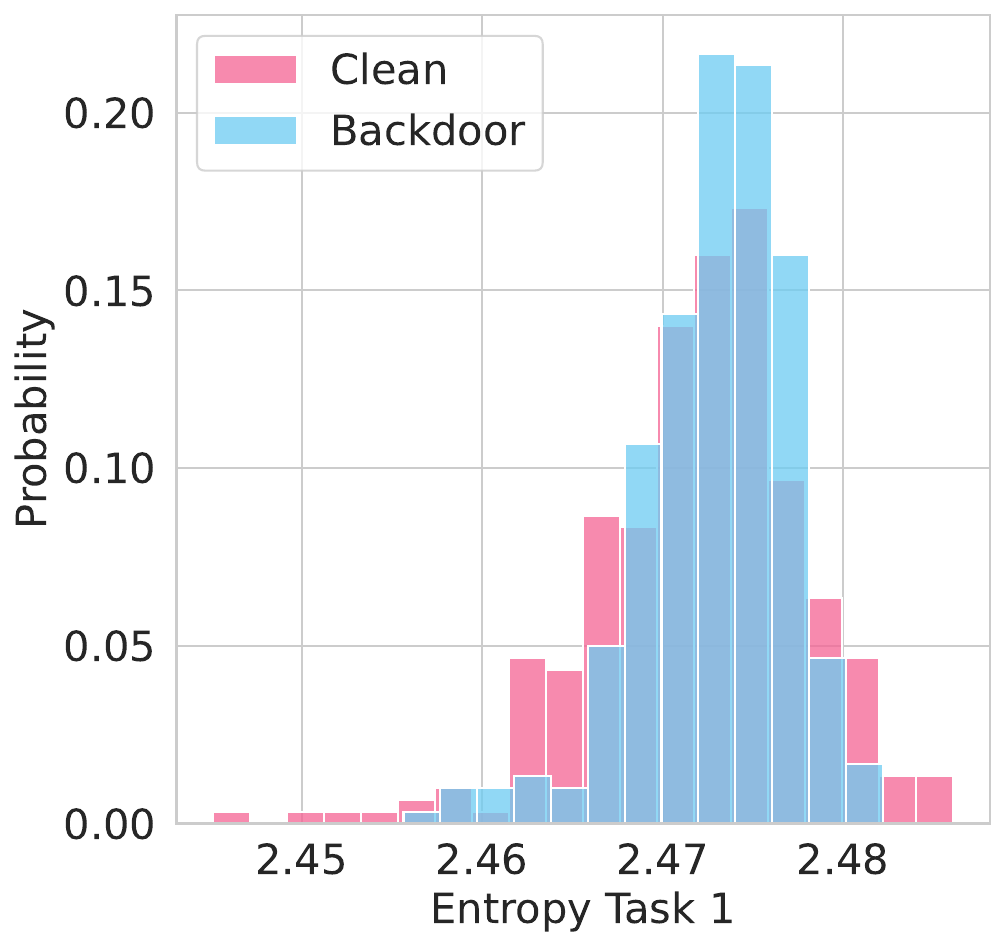}
     \end{subfigure}
     \hfill
     \begin{subfigure}[b]{0.19\textwidth}
         \centering
         \includegraphics[width=\textwidth]{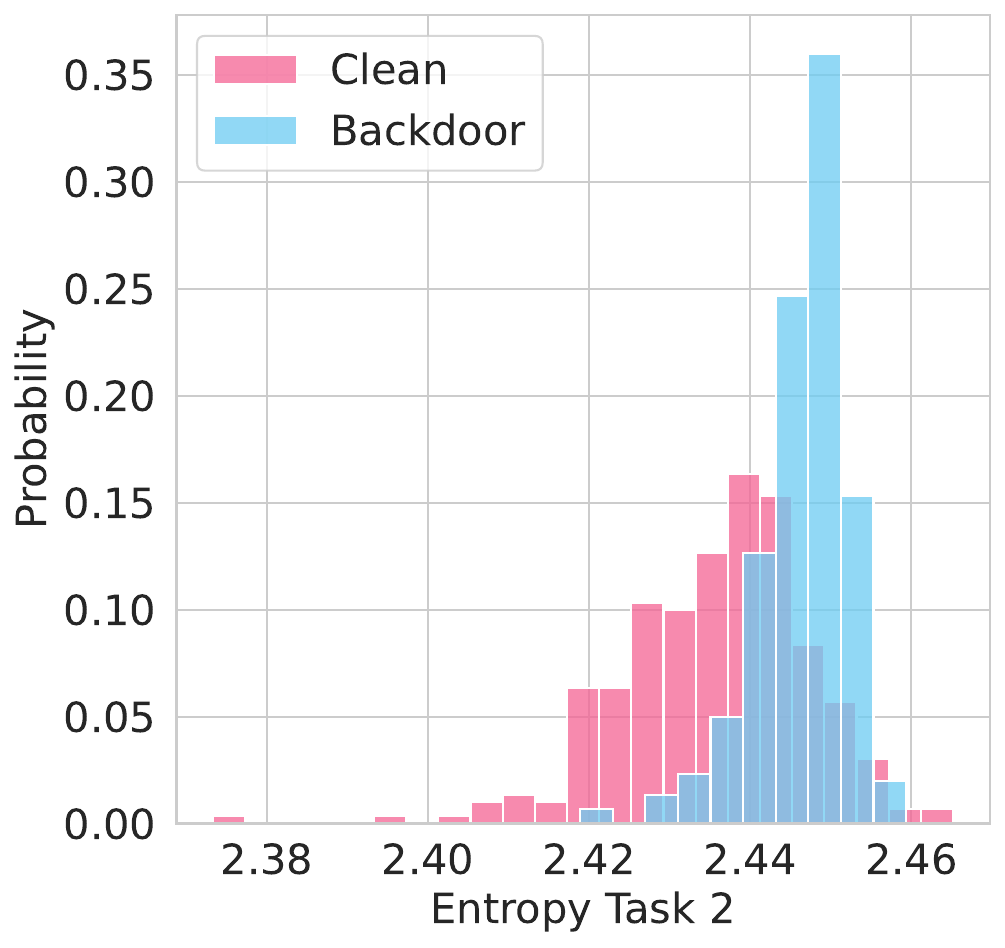}
     \end{subfigure}
     \hfill
     \begin{subfigure}[b]{0.19\textwidth}
         \centering
         \includegraphics[width=\textwidth]{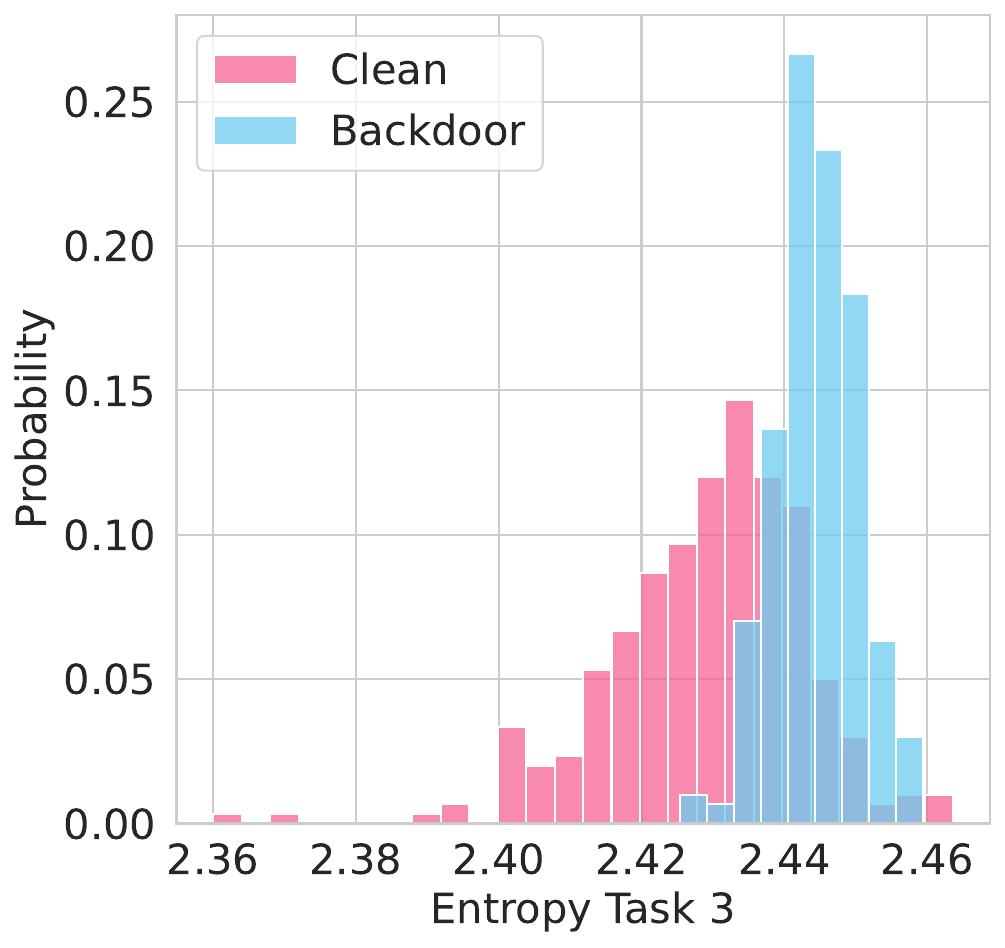}
     \end{subfigure}
     \hfill
     \begin{subfigure}[b]{0.19\textwidth}
         \centering
         \includegraphics[width=\textwidth]{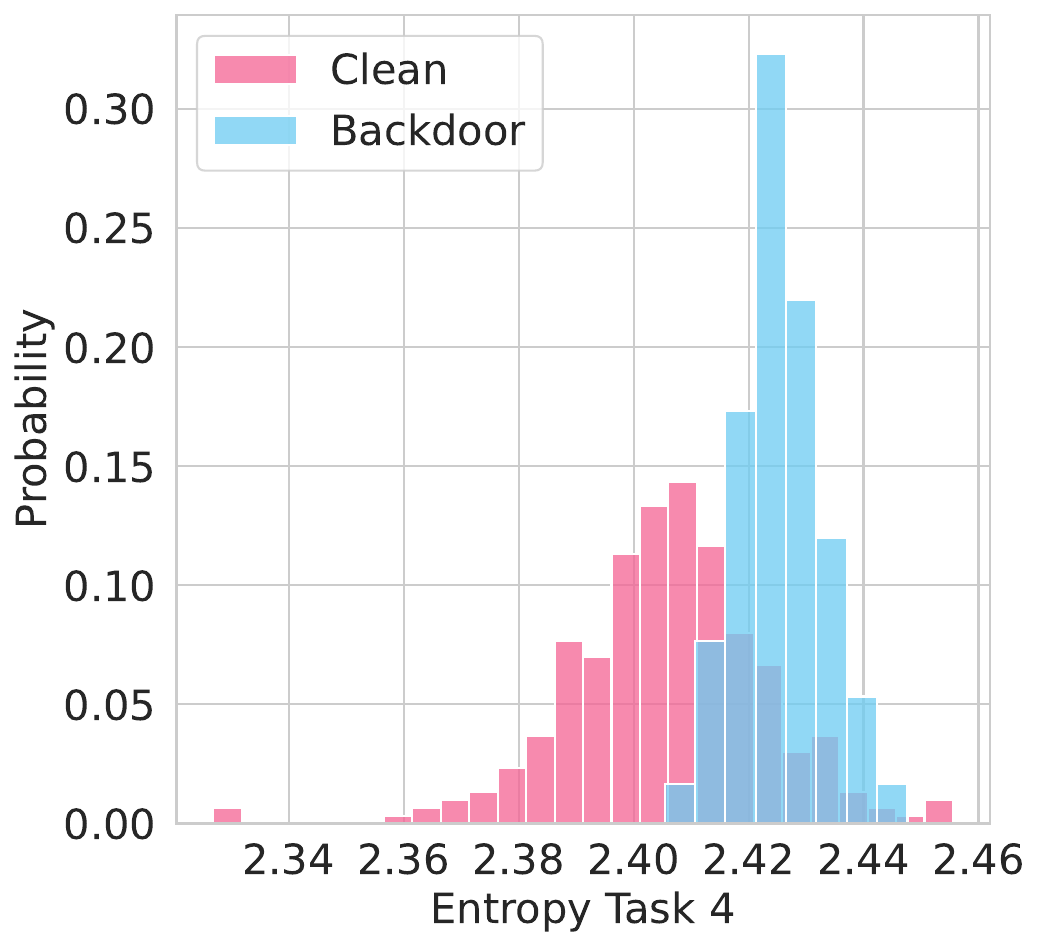}
     \end{subfigure}
     \hfill
     \begin{subfigure}[b]{0.19\textwidth}
         \centering
         \includegraphics[width=\textwidth]{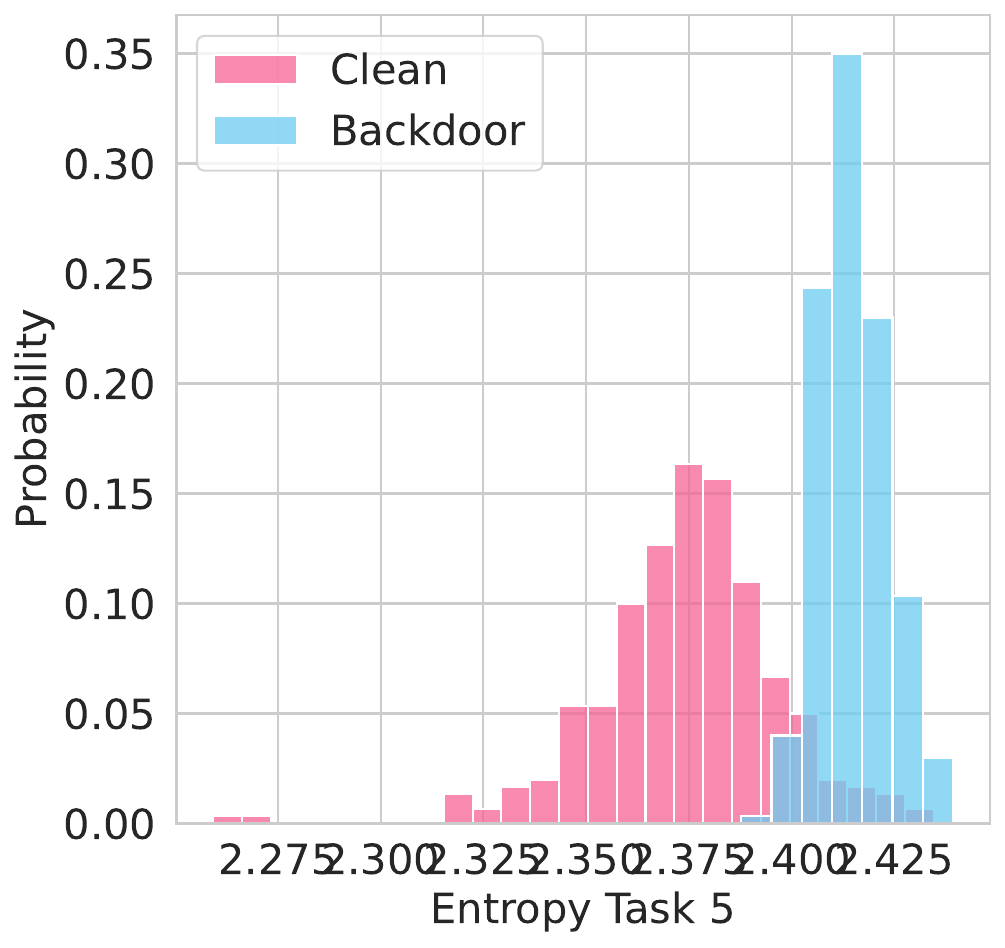}
     \end{subfigure}
     \hfill

     \begin{subfigure}[b]{0.19\textwidth}
         \centering
         \includegraphics[width=\textwidth]{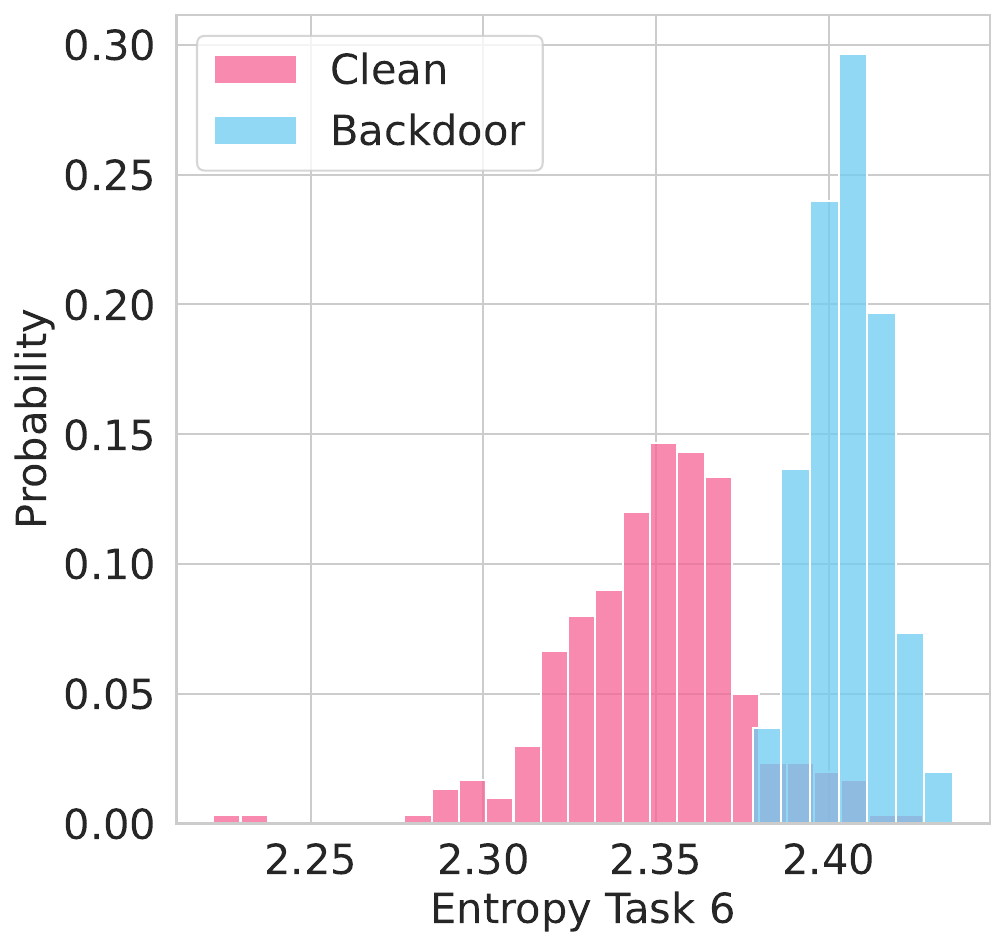}
     \end{subfigure}
     \hfill
     \begin{subfigure}[b]{0.19\textwidth}
         \centering
         \includegraphics[width=\textwidth]{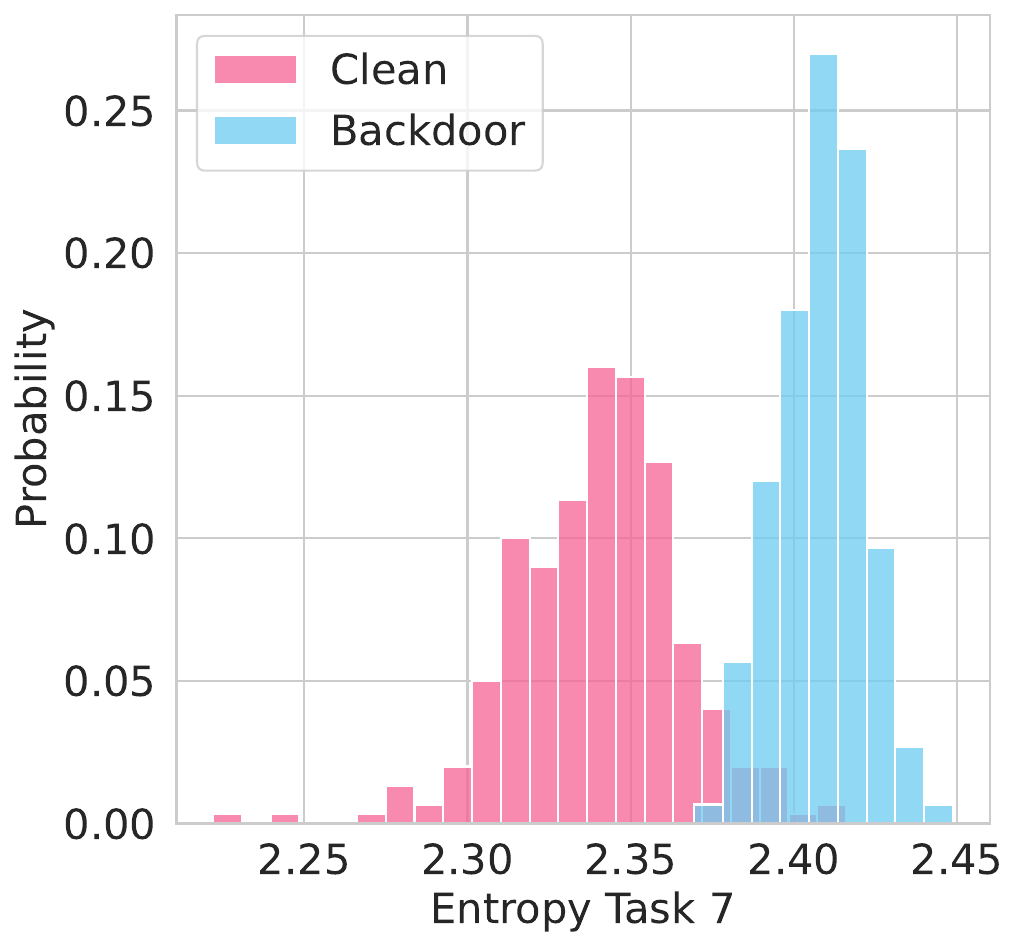}
     \end{subfigure}
     \hfill
     \begin{subfigure}[b]{0.19\textwidth}
         \centering
         \includegraphics[width=\textwidth]{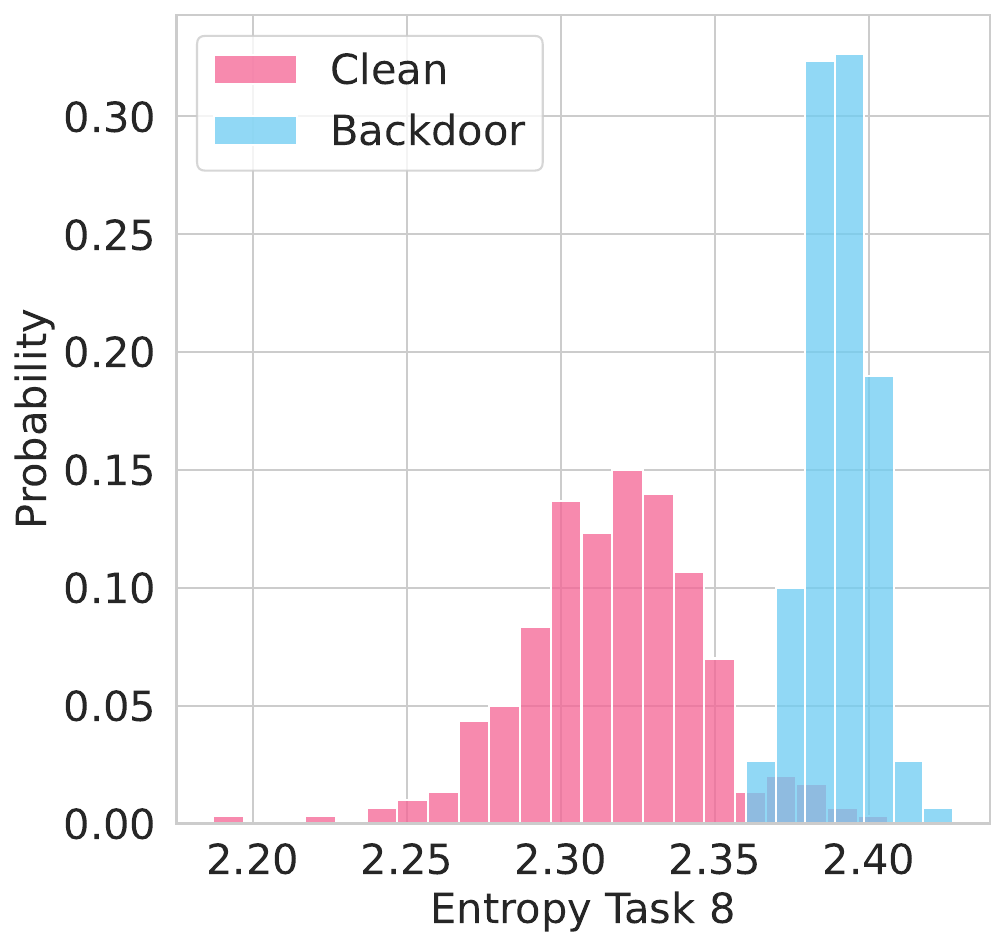}
     \end{subfigure}
     \hfill
     \begin{subfigure}[b]{0.19\textwidth}
         \centering
         \includegraphics[width=\textwidth]{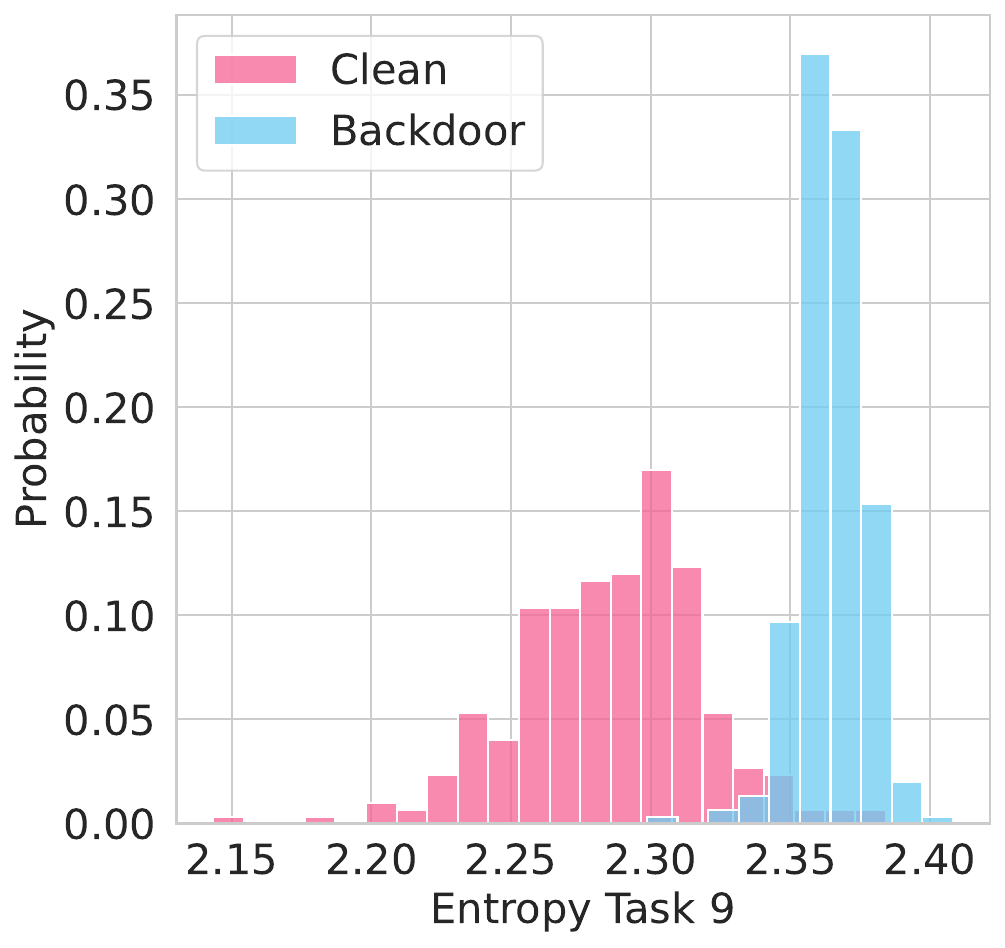}
     \end{subfigure}
     \hfill
     \begin{subfigure}[b]{0.19\textwidth}
         \centering
         \includegraphics[width=\textwidth]{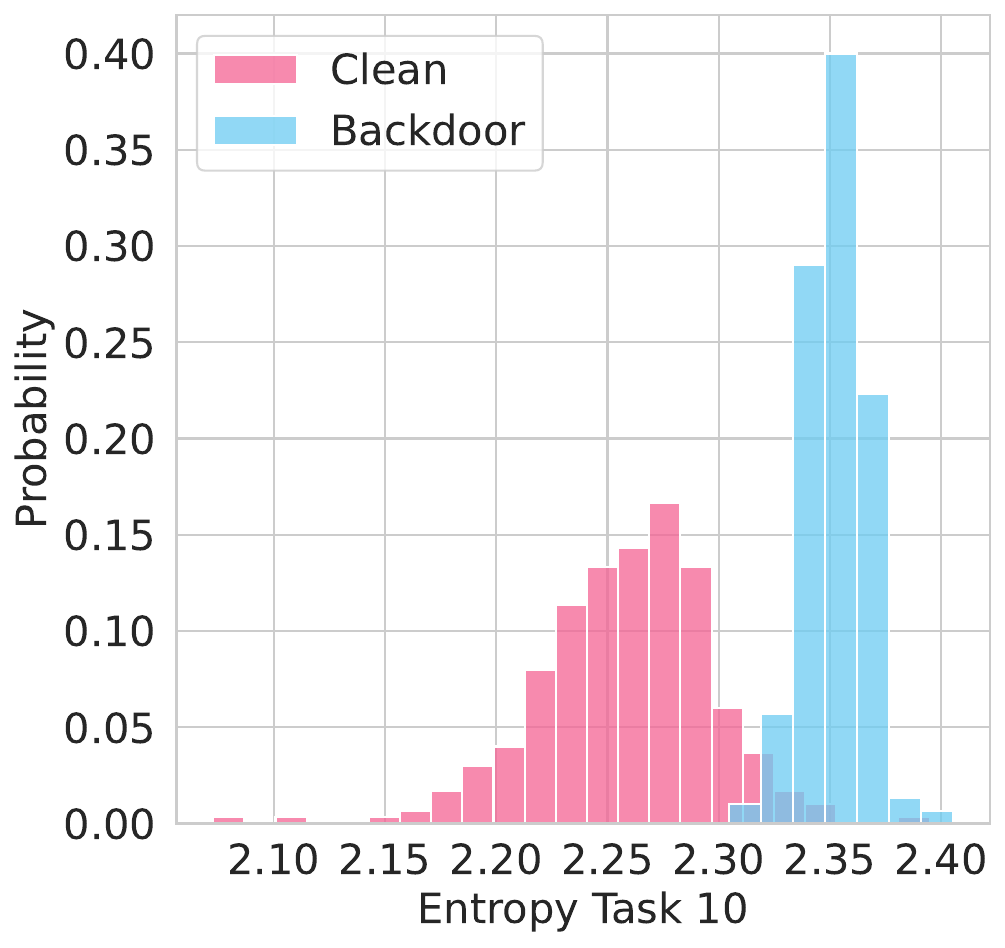}
     \end{subfigure}
     \hfill
        \caption{Comparison of AOP against STRIP. The results are visualized based on the attacked L2P on the 10-Split-ImageNet-R dataset.}
        \label{fig:strip}
\end{figure*}

\paragraph{Backdoor framework}

Our surrogate learner adopts the same settings as L2P. In the initial stage, training spans 5 epochs. Stage 2 focuses on trigger optimization, utilizing RAdam optimizer for 100 epochs with a learning rate of $0.01$. Stage 3 follows a training setting akin to stage 1. Subsequently, we initiate the dynamic stages, where the surrogate learner undergoes an update for one epoch after every 20 rounds of trigger optimization. This dynamic stage iterates for 10 rounds during attacks on Split-ImageNet-R and 5 rounds for Split-CUB200. 
For Split-ImageNet-R, the training times for stages (1) and (3) are both 2 hours, stage (2) takes 0.2 hours, and stage (4) takes 8 hours. For Split-CUB200, the training times for the four stages are 2 hours, 0.1 hours, 2 hours, and 5 hours, respectively.

\section{Additional Experiments}
\label{section:additional_experiments}

\subsection{Further discussion on AOP}

In this Appendix, we discuss the differences in ASR when using AOP to backdoor prompt-based continual learners. As shown in Table 2 of the main manuscript, in most experiments, L2P and PGP achieve the highest ASR, followed by DualPrompt and DualPrompt-PGP.

Firstly, our surrogate prompt uses the same prompt techniques and objectives as L2P, which explains its highest performance. DualPrompt introduces shared-task prompts, which might affect the ASR when updated with new classes. Additionally, unlike L2P, DualPrompt uses prefix tuning, which could cause the slight decrease in ASR. However, the ASR of DualPrompt remains higher than $96\%$, highlighting the potential for backdoor transfer between different prompt techniques. The two versions of PGP achieve performance similar to the original ones, as PGP focuses only on the update direction of prompts.

Compared to the above four versions, HiDe and CODA-Prompt show lower performance. The lower ASR of HiDe might result from using iBOT-1K as the pre-trained model for HiDe, which differs from the other learners and our surrogate learner. As prompting serves as label mapping, different source datasets might influence the mapping and thus the backdoor performance. Lastly, CODA suffers from the lowest ASR and the highest standard deviation. This is due to CODA's prompt selection mechanism, which uses an attention mechanism to get the weighted summation of all prompts, differing from the other methods.

\subsection{Additional comparison between AOP and baseline}

Narcissus \cite{zeng2022narcissus} also assumes that the attacker only has access to target data. They employ a public dataset as a surrogate dataset and optimize the trigger using the clean surrogate dataset. Our work is motivated by Narcissus, we extend the surrogate dataset in the context of prompting and exploit the label mapping property. Additionally, we employ dynamic stages and adopt BCE to prevent adversarial noise.

In Table 1 of the main manuscript, we compare AOP and Narcissus, showing that Narcissus experiences catastrophic forgetting. To provide further discussion, in Figure \ref{fig: asr_flow}, we visualize the ASR flow for each task between our AOP and Narcissus. We trained Narcissus using the same dataset and the same number of epochs as in stages (1) and (2) of our AOP. As visualized in Figure \ref{fig: asr_flow}, although Narcissus initially achieves high performance, it tends to experience catastrophic forgetting over time. Consequently, the performance gap between AOP and Narcissus increases as the training process continues.

\begin{table*}
  \caption{ACC and ASR of AOP on L2P with 10-Split-ImageNet-R when applying Vanilla FT and FST as the defense methods. Here, $\alpha$ represents the weight of the feature shifting regularization, and $N$ denotes the number of samples saved for finetuning.}
  \label{table:fst}
  \centering
  \begin{tabular}{lllcccc}
    \toprule
 & & & \multicolumn{2}{c}{$N=600$} & \multicolumn{2}{c}{$N=1200$}  \\
 & & & \multicolumn{2}{c}{$2.5\%$} & \multicolumn{2}{c}{$5\%$}  \\
 \cmidrule(r){4-5} \cmidrule(r){6-7} 
& & & ACC & ASR & ACC & ASR \\
\midrule
\multirow{2}{*}{Vanilla FT} & & $\#$ epochs = 10 & $61.25 $& $99.55$ & $67.24 $& $98.96$ \\
& & $\#$ epochs = 20 & $65.56 $& $99.07$ & $66.64 $& $97.02$ \\
\midrule
\multirow{3}{*}{FST} & $\alpha=2e-5$ & $\#$ epochs = 10 & $41.78$ & $0.00$ & $56.87$ & $0.00$ \\
 & $\alpha=2e-5$ & $\#$ epochs = 20 & $38.88$ & $0.00$ & $53.75$ & $0.00$ \\
\cmidrule(r){2-7}
 & $\alpha=2e-4$ & $\#$ epochs = 10  & $40.53$ & $0.0$ & $55.31$ & $0.0$ \\

    \bottomrule
  \end{tabular}
\end{table*}

\begin{table*}
  \caption{Backdoor performance when varying poison rates on 10-Split-ImageNet-R. $P$ denotes the number of poisoned images during training and $\gamma$ is the corresponding poisoning rate.}
  \label{table:poison_rate}
  \centering
  \begin{tabular}{lccccc}
    \toprule
& $P=0$ & $P=2$ & $P=5$ & $P=25$ & $P=100$ \\
& $\gamma=0\%$ & $\gamma=0.01\%$ & $\gamma=0.02\%$ & $\gamma=0.1\%$ & $\gamma=0.5\%$ \\
\midrule
L2P & $0.00$ & $13.76$ & $91.86$ & $99.56$ & $99.99$\\
L2P-PGP & $0.00$ & $10.08$ & $90.77$ & $99.36$ & $99.94$\\
    \bottomrule
  \end{tabular}
\end{table*}

\subsection{Defenses}

In this Appendix, we evaluate the robustness of AOP against several popular defenses, namely Neural Cleanse, STRIP, and FST.

\paragraph{Neural Cleanse} Neural Cleanse \cite{8835365} is a widely used model defense. Specifically, for each class, Neural Cleanse optimizes a trigger that induces all data to be misclassified to the target class. It then detects backdoor models by checking for abnormally small patterns among the optimized triggers using the Anomaly Index with a flag threshold of 2. We experimented with Neural Cleanse on 10-Split-ImageNet-R using checkpoint models from tasks 1, 5, and 10. AOP successfully passed Neural Cleanse as in Figure \ref{fig: neural_cleanse}.

\paragraph{STRIP} STRIP \cite{gao2019strip} is a popular test-time defense method. Given the model and a suspicious input, STRIP perturbs the input using a set of clean images from different classes and records the prediction entropy over the perturbed images. STRIP flags images as poisoned if the predictions are consistent, indicated by low entropy. We visualized the entropy of our AOP on ImageNet-R using checkpoint models from tasks 1 to 10 in Figure \ref{fig:strip}, and observed that our backdoored models exhibited a similar entropy range to benign ones, thereby passing the STRIP test.

\paragraph{Feature Shift Tuning} We evaluated AOP against a robust fine-tuning-based defense method, FST \cite{min2023towards} in Table \ref{table:fst}. FST operates by storing a small amount of clean data to fine-tune the model, reinitializing the classifier weights, and encouraging deviation from the originally compromised weights. We report the performance of FST with respect to different fine-tune data ratios as in the original paper ($2\%$ and $5\%$) and varying weights on the deviation regularizer.

We found that FST was successful in mitigating AOP, confirming its effectiveness in addressing backdoor knowledge. However, we observed that reinitializing the classifier weights results in significant forgetting, causing a considerable drop in accuracy. Thus, FST is impractical because it severely hurts the utility of the continual model while lacking verification of whether an attack exists. Furthermore, it is essential to note that FST conflicts with our data privacy prioritization scenario, as it requires storing data from all tasks.

We also evaluate Vanilla Fine-tuning (Vanilla FT) as a potential defense against AOP. As shown in Table \ref{table:fst}, while Vanilla FT preserves accuracy on clean samples, it fails to effectively mitigate the impact of AOP.

We hope our findings will inspire the development of strong defense methods compatible with multi-data supplier scenarios while upholding data privacy in continual learning.

\paragraph{Discussion on potential defenses} As observed in Figures 2a and 2b, poisoned samples consistently exhibit queries for specific prompt IDs, while clean samples demonstrate a more balanced distribution in prompt frequency selection. Consequently, potential defenses against AOP may involve monitoring the frequency selections of test samples during inference. A backdoor flag can be raised if biases in prompt selection frequencies are observed in suspected input samples. Furthermore, drawing inspiration from Fine-Pruning techniques \cite{sha2022finetuning}, which prune inactive neurons when predicting clean images, one could extend this approach to Prompt-Pruning, effectively eliminating inactive prompts.


\subsection{Sensitivity to poisoning rates}

We validate the sensitivity of AOP with respect to varying poisoning rates. We emphasize that this factor is particularly crucial in the context of backdooring CL, where the adversary only has access to the target class data—a small proportion of the overall dataset. Therefore, maintaining backdoor effectiveness with a low poisoning rate is essential.  Our AOP demonstrates favorable performance, achieving over $90\%$ accuracy even with a poisoning rate as low as $0.01\%$. This highlights the efficacy of our method in scenarios with minimal poisoning.

\section{Broader Impacts}
\label{appendix: broader impacts}

Our research contributes to the research community and AI systems by exploring the potentiality of targeted backdoor attacks in continual learning settings. By shedding light on the capabilities of such attacks, we heighten awareness about the backdoor threat, especially in private multi-data supplier scenarios. This heightened awareness encourages looking for potential protection and defenses against backdoor manipulation, a crucial key in enhancing the safety and trustworthiness of AI systems. 

Nonetheless, it is essential to acknowledge that our findings could inadvertently provide insights for attackers seeking to exploit continual learners with backdoors. Nevertheless, we believe that strong and efficient defense mechanisms will emerge to safeguard continual learners against such threats. Consequently, the positive impact of our research outweighs potential negative repercussions.

\end{document}

%% file: macro_commands.tex
\usepackage{algorithm2e}

\RestyleAlgo{ruled}

\SetKwComment{Comment}{/*}{/*}
\SetKwInput{KwInput}{Input}
\SetKwInput{KwOutput}{Output}
\SetKwInput{KwParameters}{Parameters}
\SetKwInput{KwReturn}{return}

\newlength\mylen
\newcommand\myinput[1]{%
  \settowidth\mylen{\KwIn{}}%
  \setlength\hangindent{\mylen}%
  \hspace*{\mylen}#1\\}

\LinesNumbered